\documentclass[11pt]{article}

\usepackage[final]{acl}

\usepackage{times}
\usepackage{latexsym}
\usepackage{amsmath}

\usepackage{tikz}
\usetikzlibrary{positioning, shapes.geometric, arrows.meta}
\usepackage{booktabs}
\usepackage{multirow}
\usepackage[T1]{fontenc}
\usepackage{tabularx}
\usepackage[utf8]{inputenc}

\usepackage{microtype}

\usepackage{inconsolata}

\usepackage{graphicx}
\usepackage{xurl} 
\title{Translate or Simplify First: An Analysis of Cross-lingual Text Simplification in English and French}

\author{
  Ido Dahan$^{1,5}$, Omer Toledano$^{1}$, Roey J. Gafter$^{3}$, Sharon Pardo$^{2,4}$,\\ {\bf Oren Tsur$^{1}$, Hila Zahavi$^{2,4}$, Elior Sulem$^{1,5}$} \\
  $^1$ Stein Faculty of Computer and Information Science,  Institute for Applied AI Research\\
  $^2$ The Simone Veil Research Centre for Contemporary European Studies\\
  $^3$ Department of Hebrew Language and Sociolinguistics\\
  $^4$ Department of Politics and Government\\
  $^5$ School of Statistics and Data Analysis\\
  Ben-Gurion University of the Negev \\
\texttt{idodah@post.bgu.ac.il,\{gafter,pardos,orentsur,hilape,eliorsu\}@bgu.ac.il}}

\begin{document}
\maketitle
\begin{abstract}
Cross-lingual text simplification (CLTS) aims to make content more accessible across languages by simultaneously addressing both linguistic complexity and translation. This study investigates the effectiveness of different prompting strategies for CLTS between English and French using large language models (LLMs). We examine five distinct prompting strategies: a direct prompt instructing the LLM to perform both translation and simplification simultaneously, two composition approaches that either translate-then-simplify or simplify-then-translate within a single prompt, and two decomposition approaches that perform the same operations in separate, consecutive prompts.
These systems are evaluated across a diverse set of five corpora of different genres (Wikipedia and medical texts) using nine LLMs spanning multiple model generations. Output quality is assessed through a multi-faceted evaluation framework comprising automatic metrics, comprehensive linguistic feature analysis, and human evaluation of simplicity and meaning preservation. Our findings reveal that while direct prompting achieves the highest BLEU scores in most settings, indicating meaning fidelity, Translate-then-Simplify approaches demonstrate the highest simplicity, as measured by the linguistic features.\footnote{Code and data are available at: \url{https://github.com/NLU-BGU/Cross-lingual-text-simplification-in-English-and-French}}
\end{abstract}

\section{Introduction}

\begin{figure}[hbt!]
    
    \centering
    \begin{tikzpicture}[
        node distance=2cm and 3cm,
        rect/.style={draw, rectangle, minimum width=2.3cm, minimum height=1.2cm, align=center, thick}, 
        arrow/.style={-Stealth, thick, shorten >=1pt, shorten <=1pt}
    ]
    \node[rect, fill=blue!50] (EC) at (0,0) {English\\Complex};        
    \node[rect, fill=red!50, right=of EC] (FC) {French\\Complex};      
    \node[rect, fill=blue!10, below=of EC] (ES) {English\\Simplified}; 
    \node[rect, fill=red!10, below=of FC] (FS) {French\\Simplified};   
    
    \draw[arrow] (EC) -- node[above, sloped] {\textbf{T}} (FC);
    \draw[arrow] (EC) -- node[left] {\textbf{S}} (ES);
    \draw[arrow] (ES) -- node[above, sloped] {\textbf{T}} (FS);
    \draw[arrow] (FC) -- node[auto=right] {\textbf{S}} (FS);
    \draw[arrow] (EC) -- node[above, sloped, pos=0.6, font=\small\bfseries, align=center] {CLTS} (FS);
    \end{tikzpicture}
    \caption{Conceptual diagram of text transformation tasks. Arrows represent the core operations: Translation (\textbf{T}), Simplification (\textbf{S}), and their combined operation, Cross-lingual text simplification (\textbf{CLTS}). }
    \label{fig:task_diagram}
\end{figure}
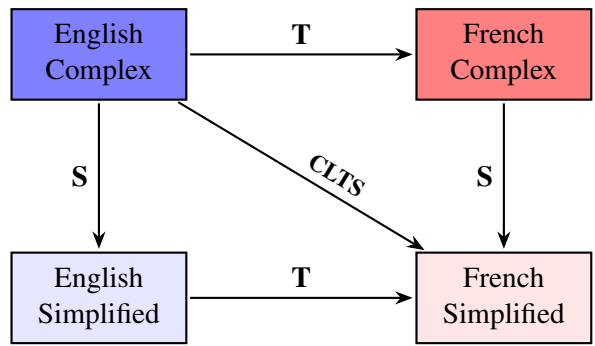

Text simplification (TS) reduces linguistic complexity while preserving meaning to improve accessibility for diverse audiences, such as language learners and individuals with reading difficulties. Traditionally, TS focuses on monolingual transformations like lexical substitution and syntactic restructuring \cite{A_survey_of_research_on_text_simplification}. However, the global demand for accessible content has created a growing need for cross-lingual text simplification (CLTS), which simultaneously translates and simplifies text across languages.

CLTS presents unique challenges that go beyond the individual tasks of translation and simplification. While machine translation systems have achieved remarkable performance in recent years, they are typically evaluated for adequacy and fluency~\cite{Leiter2023TowardsEE} rather than for accessibility to a target audience, and standard MT models do not produce simplified output by default~\cite{alva-manchego-shardlow-2022-towards}. Similarly, monolingual text simplification systems excel at reducing complexity within a single language but do not address the needs of multilingual audiences. The intersection of these two tasks requires careful consideration of how simplification strategies interact with translation processes, as linguistic complexity manifests differently across languages, for instance through their differing grammatical structures~\cite{Grammatical_complexity_in_cross_linguistic_perspective}.

CLTS can be conceptualized as a direct transformation from complex text in one language to simplified text in another. However, this single-step operation can alternatively be decomposed into sequential operations: either translating first then simplifying (T→S), or simplifying first then translating (S→T). Crucially, while Figure~\ref{fig:task_diagram} depicts these three paths as distinct routes to the same destination, they do not necessarily produce equivalent outputs. Each sequence of operations introduces different transformations, potential error propagation patterns, and linguistic trade-offs that may significantly impact the quality and characteristics of the final simplified text. This complexity is further compounded by the phenomenon of \emph{translationese} - the distinct linguistic patterns that emerge in translated texts, which often include implicit simplification effects such as reduced lexical variety~\cite{On_the_features_of_translationese}.

Traditional automatic TS approaches rely on parallel complex--simple training data, which is scarce outside English \cite{alva-manchego-etal-2020-data,ryan-etal-2023-revisiting}. Large language models (LLMs) offer zero-shot capabilities in both translation and simplification, but how best to combine the two operations with LLMs remains an open question.

Recent advances in prompt engineering have shown that the way tasks are presented to LLMs can dramatically influence their performance. Showing models in context how to compose basic skills improves their compositional generalization~\cite{chen-etal-2024-skills}. Similarly, decomposition approaches that explicitly separate different sub-tasks may offer advantages over single-step approaches~\cite{khot2023decomposedpromptingmodularapproach}. However, the effectiveness of these different prompting strategies for CLTS has not been systematically investigated.

\begin{table*}[!t]
\centering
\definecolor{lightgray}{gray}{0.95}
\definecolor{darkblue}{RGB}{50,80,120}

\setlength{\fboxrule}{0.1pt}
\setlength{\fboxsep}{12pt}

\fbox{%
\begin{minipage}{0.95\linewidth}
    \normalsize\textbf{ \textcolor{darkblue}{Original Text (English):}}\\[0.3em]
    A few animals have chromatic response, changing color in changing environments, either seasonally (ermine, snowshoe hare) or far more rapidly with chromatophores in their integument (the cephalopod family).

    \begin{center}
    \small
    \begin{tabular}{p{0.48\linewidth}p{0.48\linewidth}}
        \textbf{\textcolor{darkblue}{French Simplification}} & \textbf{\textcolor{darkblue}{English Translation}} \\
        \hline
        \\[-0.8em]
        
        \textbf{Direct Simplification:} & \textbf{Translation:} \\[0.1em]
        Quelques animaux changent de couleur pour s'adapter à leur environnement, soit au fil des saisons (hermine, lièvre d'Amérique), soit plus rapidement grâce à des cellules spéciales dans leur peau (les céphalopodes).
        &
        Some animals change color to adapt to their environment, either with the seasons (ermine, snowshoe hare), or more quickly thanks to special cells in their skin (cephalopods). \\
        \\[0.3em]
        \hline
        \\[-0.8em]
        
        \textbf{$T \rightarrow S$ Composition:} & \textbf{Translation:} \\[0.1em]
        Certains animaux changent de couleur pour s'adapter à leur environnement. Ce changement peut être saisonnier (comme chez l'hermine ou le lièvre d'Amérique) ou très rapide grâce à des cellules spéciales dans leur peau (comme chez les calmars ou les poulpes).
        &
        Some animals change color to adapt to their environment. This change can be seasonal (as in ermine or snowshoe hare) or very rapid thanks to special cells in their skin (as in squid or octopus). \\
        \\[0.3em]
        \hline
        \\[-0.8em]
        
        \textbf{$S \rightarrow T$ Decomposition:} & \textbf{Translation:} \\[0.1em]
        Certains animaux changent de couleur pour se fondre dans leur environnement, soit saisonnièrement (comme l'hermine), soit rapidement grâce à des cellules cutanées spéciales (comme les pieuvres).
        &
        Some animals change color to blend into their surroundings, either seasonally (like stoats) or rapidly using special skin cells (like octopuses). \\
    \end{tabular}
    \end{center}
\end{minipage}%
}

\footnotesize

\caption{Original English text from the ASSET corpus with multiple French simplifications generated by Gemini 2.5 Flash-Lite using 3 different prompting techniques. Abbreviations: T = Translate, S = Simplify.}
\label{tab:original_with_simplifications}
\end{table*}

Because automatic metrics capture simplification quality only partially and may favor different strategies, we complement them with a linguistic-feature analysis and with human annotation of meaning preservation and simplicity.

We study bidirectional CLTS, from English to French and from French to English, which allows us to investigate directional asymmetries in simplification strategies and their effectiveness. English--French is a deliberately controlled starting point: two high-resource languages with strong reference data, which isolates the effect of prompting structure from translation-quality confounds and gives access to a diverse range of datasets. Our conclusions should accordingly be read within this setting. Table~\ref{tab:original_with_simplifications} illustrates example outputs from three of these approaches applied to an English source text from the ASSET corpus \citep{alva-manchego-etal-2020-asset}.

This study makes two contributions. First, to the best of our knowledge, it is the first study to disentangle the five distinct ways of combining translation and simplification (direct, T$\rightarrow$S and S$\rightarrow$T composition, T$\rightarrow$S and S$\rightarrow$T decomposition) rather than treating CLTS as a black box. We compare them along three complementary axes (automatic metrics, a cross-lingual adaptation of the BATS linguistic features~\cite{kreutz-etal-2024-bats}, and human judgment), on five corpora from two domains, in both translation directions, and with nine LLMs. We show that operation order and single- vs.\ multi-prompt execution yield systematically different outputs: direct prompting is the most faithful but simplifies least, translate-then-simplify strategies produce the most readable outputs, and decomposition introduces a structurally specific risk of propagated errors. These findings offer transferable insights for designing CLTS pipelines. Second, we provide a reusable evaluation protocol for CLTS, combining multiple datasets with multi-faceted evaluations adapted to the task, which can be re-applied to additional language pairs.
\section{Related Work}
\subsection{Cross-lingual Simplification}

While monolingual text simplification has seen significant advancements, cross-lingual simplification remains an understudied area. Early work by \citet{mallinson-etal-2020-zero} demonstrated the feasibility of zero-shot knowledge transfer from English to German.
More recently, the MultiCochrane dataset \cite{joseph-etal-2023-multilingual} provided parallel sentence-aligned data for medical simplification across English, French, Spanish, and Farsi, facilitating research on how simplification strategies transfer across specific language pairs. This research is further supported by \citet{ryan-etal-2023-revisiting}, who utilize a 12-language benchmark to demonstrate effective zero-shot cross-lingual transfer of simplification models.
\citet{joseph-etal-2023-multilingual} further compare direct multilingual simplification with a simplify-then-translate pipeline (GPT-3 simplification followed by machine translation) and find the pipeline fairly strong relative to direct multilingual simplification.
Closest to our setting, \citet{shardlow-alva-manchego-2022-simple} release Simple TICO-19, a dataset for joint translation and simplification of COVID-19 texts (English--Spanish), and \citet{alva-manchego-shardlow-2022-towards} show that standard MT models do not generate simplified translations by default, leaving translate-then-simplify and simplify-then-translate pipelines, as well as joint models, to future work. 
These works show that cross-lingual simplification is feasible, but they do not compare both operation orders, nor single- vs.\ multi-prompt execution, with current LLMs.
Our work complements these directions by systematically comparing joint and sequential strategies, and studying how the ordering of the two operations and their execution (within one prompt or across prompts) shape the output.

\subsection{LLMs for (Cross-lingual) Text Generation}

Large pre-trained models can perform many NLP tasks zero-shot through prompting \citep{brown2020languagemodelsfewshotlearners}, and eliciting intermediate reasoning steps further improves performance on complex tasks \citep{wei2023chainofthoughtpromptingelicitsreasoning}. For cross-lingual tasks, \citet{wang-etal-2023-zero} proposed prompting strategies for zero-shot cross-lingual summarization, which we adapt by replacing summarization instructions with simplification objectives.

\subsection{Evaluation of Text Simplification}
Simplification involves trade-offs between meaning preservation, readability, and linguistic adequacy, which makes its evaluation challenging~\cite{alva-manchego-etal-2021-un}.

\subsubsection{Automatic Evaluation of Simplification}

The evaluation of text simplification has traditionally relied on machine translation metrics such as BLEU~\cite{papineni-etal-2002-bleu}. SARI~\cite{xu-etal-2016-optimizing} was introduced to explicitly reward additions, deletions, and retentions, and is now widely used for simplification evaluation. \citet{xu-etal-2016-optimizing} found that BLEU correlates reasonably with grammaticality and meaning preservation but fails to capture simplicity; \citet{sulem-etal-2018-bleu} further showed that BLEU often correlates negatively with simplicity, penalizing legitimate simplification operations such as sentence splitting, and that its correlation with grammaticality and meaning preservation is low or absent when splitting is involved. More recently, semantic similarity metrics such as BERTScore~\cite{zhang2020bertscoreevaluatingtextgeneration} have been adopted to better capture meaning preservation in text generation tasks. Surveys highlight that no single metric fully captures simplification quality and recommend combining multiple measures~\cite{grabar-saggion-2022-evaluation,alva-manchego-etal-2021-un}.

\subsubsection{Feature-Based Evaluation of Simplicity}

Several studies have emphasized the role of linguistic features in assessing text simplicity. ~\citet{vajjala2016readabilitybasedsentencerankingevaluating} showed that word-level and syntactic features help rank sentences by readability and distinguish simplified from original sentences. Recent work has expanded this direction by combining transformer-based models with diverse linguistic features, to improve the explainability of simplification systems \cite{qiao-etal-2022-psycho}. \citet{kreutz-etal-2024-bats} introduced BATS (BenchmArking Text Simplicity), a comprehensive framework that implements 37 literature-derived features across multiple dimensions. We adopt the majority of BATS features in our evaluation framework, applying them to assess cross-lingual simplification, a novel extension beyond the original monolingual focus of BATS.

\subsubsection{Human Evaluation of Simplification}

Despite the efficiency of linguistic features, human judgments remain the reference against which automatic simplification metrics are assessed \cite{alva-manchego-etal-2021-un}, and reading-comprehension studies with target readers are considered the most reliable way to measure accessibility \cite{yaneva-etal-2016-evaluating}. Traditional human evaluation frameworks focus on three primary dimensions: \emph{meaning preservation} (adequacy), \emph{grammaticality} (fluency), and \emph{simplicity} \cite{xu-etal-2016-optimizing,nisioi-etal-2017-exploring,sulem-etal-2018-simple}. However, obtaining reliable human judgments in simplification is challenging due to the inherent subjectivity of ``simplicity'', which can vary significantly depending on the target audience's profile \cite{grabar-saggion-2022-evaluation}. In the cross-lingual context, human judgment is particularly vital to account for the interplay between translation accuracy and readability, which automatic metrics may overlook. We therefore incorporate human evaluation to validate whether improvements in automatic scores and linguistic features translate to genuine gains in accessibility and fidelity for bilingual audiences.

\section{Prompting Strategies in CLTS}
\label{sec:prompting}

CLTS combines two transformations: translation and simplification. Strategies for combining them differ along three dimensions: (1) whether the two operations are performed jointly in a single instruction (\textbf{Direct}) or sequentially; (2) if sequentially, which comes first: \textbf{Translate-then-Simplify} (T$\rightarrow$S) or \textbf{Simplify-then-Translate} (S$\rightarrow$T); and (3) whether the sequential operations are specified within a single prompt (\textbf{Composition}) or executed in separate consecutive prompts, the output of the first serving as input to the second (\textbf{Decomposition}). This yields five strategies, all explored in this work: Direct, T$\rightarrow$S and S$\rightarrow$T Composition, and T$\rightarrow$S and S$\rightarrow$T Decomposition (prompts in Section~\ref{sec:prompts}).

\begin{table*}[t]
\centering
\small
\begin{tabular}{p{3.5cm}|p{11.5cm}}
\toprule
\textbf{Feature Category} & \textbf{Features} \\
\midrule
Lexical Features & lexical richness; infrequent words ratio; long words ratio; content words ratio; average word length \\
\hline
Syntactic and Structural Features & words before main verb; noun phrases ratio; relative clauses ratio; appositions ratio; conditional clauses ratio; conjunctions ratio; passive voice ratio; syntactic tree depth; sentences number; words per sentence; short sentences ratio \\
\hline
Readability Features & Flesch Reading Ease; Flesch Kincaid Grade; syllables ratio \\
\hline
Named Entity Features & max same entity distance; unique entities; unique entities average; avg same entity distance; entity to token ratio; unique entities to total num of entities; consecutive entity distance \\
\hline
Grammatical Features & modifiers ratio; negations ratio; past tense verbs; punctuation ratio; third person pronouns ratio \\
\bottomrule
\end{tabular}
\caption{Comprehensive linguistic features used for CLTS analysis.}
\label{tab:linguistic_features}
 \vspace{-0.5em}
\end{table*}

Each choice involves \textbf{trade-offs}: (i) Parallel vs. Sequential processing (joint optimization vs. clear decomposition with error propagation); (ii) Simplifying First (eases translation but loses information early) vs. Translating First (preserves information but complicates target-language simplification); (iii) Single vs. Multiple Prompts (better integration vs. task focus without coordination).




Our systematic evaluation addresses key questions about these strategic choices:
(1) Which prompting strategy yields the best performance across different evaluation metrics (BLEU, SARI, semantic similarity)?
(2) How do different strategies differ in the specific linguistic transformations they perform (features in Table~\ref{tab:linguistic_features})?
(3) Does human evaluation confirm the trade-off between meaning preservation and simplicity across different prompting strategies? 

\section{Experimental Setup}

\subsection{Datasets}

We evaluate our approaches on five corpora representing diverse domains and language pairs. Consistent with standard benchmarking in text simplification, all datasets used in this study are sentence-based, where each instance consists of a single complex sentence and its simplified counterpart(s). Corpus sizes are given in Table~\ref{corpus_statistics_tab}.

\begin{table}[t]
\centering
\small
\begin{tabular}{@{}llrr@{}}
\toprule
\textbf{Corpus} & \textbf{Dir.} & \textbf{Train} & \textbf{Test} \\
\midrule
ASSET         & EN$\rightarrow$FR & 20,000  & 359   \\
WikiAuto      & EN$\rightarrow$FR & 576,126 & 4,690 \\
MultiCochrane & EN$\rightarrow$FR & 28,998  & 264   \\
WikiLarge-FR  & FR$\rightarrow$EN & 264,258 & 1,067 \\
CLEAR         & FR$\rightarrow$EN & 4,110   & 389   \\
\bottomrule
\end{tabular}
\caption{Corpus sizes (sentence pairs) and CLTS direction. Training splits are used only to fine-tune the mT5 baseline; the LLMs are prompted zero-shot on the test splits.}
\label{corpus_statistics_tab}
\end{table}

\paragraph{ASSET:} An English Wikipedia dataset with multiple human simplifications crowdsourced for each sentence
~\cite{alva-manchego-etal-2020-asset}.

\paragraph{WIKI-AUTO:} A large-scale English dataset automatically aligned from English Wikipedia and Simple English Wikipedia~\cite{jiang-etal-2020-neural}.

\paragraph{MultiCochrane:} The first multilingual, sentence-aligned medical text simplification dataset, pairing complex English texts with simplified texts in English, Spanish, French, and Farsi~\cite{joseph-etal-2023-multilingual}.

\paragraph{CLEAR:} A French medical corpus of comparable texts from sources like encyclopedias and drug leaflets, with a subset of manually aligned parallel sentences~\cite{grabar-cardon-2018-clear}.

\paragraph{WikiLarge-Fr:} A French version of WikiLarge~\cite{zhang-lapata-2017-sentence}, automatically translated from English~\cite{cardon-grabar-2020-french}.
\begin{table*}[htbp]
\centering
\footnotesize
\setlength{\tabcolsep}{2.5pt} 
\resizebox{\textwidth}{!}{%
\begin{tabular}{@{}ll cc c cc c cc c cc c cc c@{}}
\toprule
 & & \multicolumn{3}{c}{\textbf{GPT-3.5}} & \multicolumn{3}{c}{\textbf{GPT-4o-m}} & \multicolumn{3}{c}{\textbf{GPT-5.4-Nano}} & \multicolumn{3}{c}{\textbf{Gemini 2.5}} & \multicolumn{3}{c}{\textbf{DeepSeek}} \\
\cmidrule(lr){3-5} \cmidrule(lr){6-8} \cmidrule(lr){9-11} \cmidrule(lr){12-14} \cmidrule(lr){15-17}
\textbf{Corpus} & \textbf{Method} & \textbf{B} & \textbf{S} & \textbf{CB} & \textbf{B} & \textbf{S} & \textbf{CB} & \textbf{B} & \textbf{S} & \textbf{CB} & \textbf{B} & \textbf{S} & \textbf{CB} & \textbf{B} & \textbf{S} & \textbf{CB} \\
\midrule
ASSET & Direct & \textbf{66.09}$^{***}$ & 39.18 & \textbf{0.84}$^{***}$ & \textbf{70.14}$^{***}$ & 39.17 & \textbf{0.84}$^{***}$ & 56.43 & 40.55 & 0.83 & \textbf{68.13}$^{***}$ & 39.71 & \textbf{0.83}$^{***}$ & 65.18 & 40.49 & 0.83 \\
 & $T \rightarrow S$ C. & 57.11 & 42.56 & 0.81 & 58.78 & 43.91 & 0.80 & 51.11 & 41.71 & 0.81 & 47.13 & 41.45 & 0.76 & 63.79 & 41.53 & 0.83 \\
 & $S \rightarrow T$ C. & 59.47 & 41.75 & 0.80 & 60.50 & 44.28 & 0.81 & 54.47 & 42.16 & 0.82 & 48.92 & 40.51 & 0.74 & 67.15 & 40.42 & 0.82 \\
 & $T \rightarrow S$ D. & 51.89 & 42.86 & 0.82 & 60.02 & 43.46 & 0.81 & 48.30 & 41.44 & 0.81 & 52.50 & 42.02 & 0.77 & 55.91 & 43.18 & 0.80 \\
 & $S \rightarrow T$ D. & 48.03 & 43.68 & 0.81 & 60.38 & 44.19 & 0.81 & 54.18 & 42.38 & 0.81 & 54.24 & 42.10 & 0.77 & 52.28 & 42.42 & 0.78 \\
\midrule
WikiAuto & Direct & \textbf{30.97}$^{***}$ & 32.24 & \textbf{0.79}$^{***}$ & \textbf{32.41}$^{***}$ & 33.28 & \textbf{0.80}$^{***}$ & \textbf{26.38}$^{***}$ & 33.91 & \textbf{0.79}$^{***}$ & \textbf{29.95}$^{***}$ & 33.24 & \textbf{0.79}$^{***}$ & 29.36 & 34.15 & 0.79 \\
 & $T \rightarrow S$ C. & 21.89 & 33.36 & 0.76 & 21.37 & 34.44 & 0.76 & 21.83 & 33.97 & 0.77 & 16.64 & 32.87 & 0.73 & 28.95 & 34.77 & 0.79 \\
 & $S \rightarrow T$ C. & 23.79 & 33.48 & 0.76 & 24.32 & 35.00 & 0.77 & 23.99 & 34.16 & 0.78 & 17.26 & 32.74 & 0.71 & 29.69 & 33.99 & 0.79 \\
 & $T \rightarrow S$ D. & 22.02 & 33.70 & 0.77 & 25.63 & 35.26 & 0.78 & 21.68 & 33.97 & 0.77 & 18.88 & 33.65 & 0.74 & 22.40 & 34.64 & 0.77 \\
 & $S \rightarrow T$ D. & 20.27 & \textbf{34.25} & 0.76 & 25.47 & 35.41 & 0.77 & 23.52 & 34.23 & 0.77 & 19.75 & 33.84 & 0.74 & 20.01 & 34.07 & 0.75 \\
\midrule
M-Coch. & Direct & 10.78 & 32.26 & 0.59 & \textbf{13.46} & 33.17 & 0.66 & 11.04 & 38.15 & 0.66 & \textbf{12.93}$^{***}$ & 34.78 & \textbf{0.66} & 11.96 & 36.77 & 0.66 \\
 & $T \rightarrow S$ C. & 10.05 & 36.43 & 0.64 & 9.54 & 38.68 & 0.64 & 9.07 & 39.14 & 0.64 & 7.19 & 38.87 & 0.61 & 11.29 & 36.60 & 0.66 \\
 & $S \rightarrow T$ C. & 12.73 & 34.56 & 0.65 & 10.10 & 38.99 & 0.65 & 10.43 & 39.22 & 0.66 & 8.26 & 39.62 & 0.61 & 12.14 & 36.51 & 0.66 \\
 & $T \rightarrow S$ D. & 9.88 & 37.93 & 0.65 & 10.45 & 37.79 & 0.65 & 8.94 & 38.95 & 0.65 & 7.92 & 39.05 & 0.62 & 9.43 & 39.77 & 0.64 \\
 & $S \rightarrow T$ D. & 9.56 & 38.93 & 0.65 & 10.63 & 39.21 & 0.66 & 9.69 & 38.66 & 0.66 & 8.89 & 40.28 & 0.63 & 8.68 & 40.27 & 0.64 \\
\bottomrule
\end{tabular}%
}
\caption{Automatic Evaluation Metrics (\textbf{English to French}). Bold: significantly better than each other strategy ($p<0.05$, Welch's $t$-test); $^{**}$: $p<0.01$; $^{***}$: $p<0.001$. Abbreviations: B=BLEU, S=SARI, CB=CamemBERTScore, C.=Composition, D.=Decomposition.}
\label{tab:english_metrics_compact}
\end{table*}

\subsection{Data Preprocessing}
\label{sec:data_preprocessing}

Current text simplification datasets are primarily designed for monolingual settings and lack the parallel structures necessary for evaluating cross-lingual simplification. To address this limitation, we used machine translation to create the required cross-lingual evaluation data for monolingual corpora. Specifically, for each original-simplified text pair in a source language, we translated both texts to the target language. The translation was performed using the \textbf{Google Translate API}. This process created the original text and reference simplified texts in the target language. \texttt{MultiCochrane} already provides cross-lingual parallel data and did not require this step. To ensure semantic consistency, we filtered out pairs whose Sentence-BERT~\cite{reimers-gurevych-2019-sentence} similarity was below 0.6, a cut-off chosen by manual inspection of pairs around the boundary: above 0.6, the core proposition was preserved; below, pairs frequently exhibited topic drift or missing clauses.

\subsection{Models}
We evaluate our strategies using nine LLMs spanning multiple model generations and encompassing both proprietary and open-source architectures. Proprietary models include \textbf{GPT-3.5-Turbo}, \textbf{GPT-4o-Mini}, \textbf{GPT-5.4-Nano}, and \textbf{Gemini 2.5 Flash-Lite}. We also use open-source models including \textbf{DeepSeek-Chat}, \textbf{Aya101} \citep{aya101-paper}, \textbf{Mistral-NeMo}, \textbf{GPT-OSS-20B}, and a fine-tuned \textbf{mT5} baseline \citep{xue2021mt5massivelymultilingualpretrained} to ensure the reproducibility of our findings and promote transparent benchmarking in the research community. Legacy models such as GPT-3.5-Turbo are included deliberately as cross-generation controls, allowing us to test whether the strategy-level patterns persist from older to more recent models. Detailed documentation and resource links for each model are provided in Appendix \ref{sec:model_docs}, with specific implementation details for the mT5 baseline detailed in Appendix \ref{sec:mt5_details}.

\section{Methodology}

\subsection{Prompting Strategies}
\label{sec:prompts}

The five strategies of Section~\ref{sec:prompting} are implemented with the following prompts (shown for French as the target language; the same structure applies when English is the target). All prompts begin with ``Please'' to keep the surface form consistent across strategies, so that performance differences can be attributed to operation order and prompt structure rather than to instruction phrasing:

\paragraph{Direct:} \textit{``Please simplify the following text in French:''}

\paragraph{T$\rightarrow$S Composition:} \textit{``Please first translate the following text to French and then simplify the translated text in French:''}

\paragraph{S$\rightarrow$T Composition:} \textit{``Please first simplify the following text and then translate the simplification to French:''}

\paragraph{T$\rightarrow$S Decomposition:} Step 1: \textit{``Please translate the following text to French:''}; Step 2: \textit{``Please simplify the following text in French:''}

\paragraph{S$\rightarrow$T Decomposition:} Step 1: \textit{``Please simplify the following text:''}; Step 2: \textit{``Please translate the following text to French:''}

All prompts are formulated in English; a robustness experiment using French-language prompts is reported in Appendix~\ref{app:french_prompt}. To test robustness to phrasing, we also re-ran all strategies with GPT-5.4-Nano under two paraphrased prompt sets that change style and information density and avoid the word ``simplify'' (Appendix~\ref{app:paraphrase}). Under both sets, Direct remains the most source-faithful strategy (highest BLEU and semantic similarity on every corpus) and T$\rightarrow$S composition the most readable (highest Flesch Reading Ease on every corpus).


\subsection{Evaluation Metrics}

We employ a comprehensive evaluation framework using multiple automatic metrics to assess different aspects of cross-lingual simplification quality:

\paragraph{BLEU~\cite{papineni-etal-2002-bleu}:} Measures n-gram overlap between generated and reference texts, providing an indication of translation quality.

\paragraph{SARI ~\cite{xu-etal-2016-optimizing}:} Evaluates simplification quality by comparing n-gram additions, deletions, and retentions against references and the source.

\paragraph{Semantic Similarity Metrics:} To capture meaning preservation across languages, we use language-specific semantic similarity scores: {\bf BERTScore}~\cite{zhang2020bertscoreevaluatingtextgeneration} for French-to-English outputs, and {\bf CamemBERTScore} (which substitutes CamemBERT embeddings for BERT \citep{Martin_2020}) for English-to-French outputs.

All reported scores are averages of sentence-level scores. We test differences between strategies with Welch's two-sample $t$-test (unequal variances) on these sentence-level scores, run separately for each combination of corpus, model, and metric. A strategy is marked as significantly best (bold in the result tables) when it outperforms each of the other four strategies at $p<0.05$; superscripts $^{**}$ and $^{***}$ indicate that the weakest of these four comparisons reaches $p<0.01$ and $p<0.001$, respectively. No correction for multiple comparisons is applied, so these markers should be read as indicative.

\subsection{Human Evaluation}

To complement our automatic evaluation, we conducted a human annotation study on the first 70 texts from each corpus using GPT-3.5-Turbo outputs. Because the strategy-level trends in the automatic and feature-based results recur across models, we restricted the study to a single model so as to cover all corpora and all five strategies; the error analysis (Section~\ref{sec:error_analysis_main}) uses the same model, keeping our two manual studies directly comparable. The annotators were five students, native speakers of French with high proficiency in English; they were not medical experts and were compensated for their work. For all corpora, annotators compared the outputs against the original source texts. Additionally, for the WikiAuto and CLEAR corpora, we extended the study to include a comparison against the translated versions of the texts. Most comparisons were reviewed by two annotators, with several instances receiving three. Participants assessed meaning preservation using 1 to 5 intensity scales to measure both the extent of information addition and the extent of information removal. Overall simplicity was rated on a -2 to 2 scale. The full guidelines appear in Appendix~\ref{app:annotation_guidelines}.

\subsection{Feature-based Analysis}

To gain deeper insights into the characteristics of successful CLTS, we extract and analyze a comprehensive set of linguistic features from the outputs of each prompting strategy. These features were designed to capture various linguistic and structural properties, which are detailed in Table~\ref{tab:linguistic_features}. A comprehensive glossary and description of each feature can be found in Appendix~\ref{features_descriptions}.

We apply t-tests to compare these linguistic features between the outputs of different prompting strategies for each model and corpus combination.

\begin{table*}[htbp]
\centering
\footnotesize
\setlength{\tabcolsep}{3pt} 
\resizebox{\textwidth}{!}{%
\begin{tabular}{@{}ll cc c cc c cc c cc c cc c@{}}
\toprule
 & & \multicolumn{3}{c}{\textbf{GPT-3.5}} & \multicolumn{3}{c}{\textbf{GPT-4o-m}} & \multicolumn{3}{c}{\textbf{GPT-5.4-Nano}} & \multicolumn{3}{c}{\textbf{Gemini 2.5}} & \multicolumn{3}{c}{\textbf{DeepSeek}} \\
\cmidrule(lr){3-5} \cmidrule(lr){6-8} \cmidrule(lr){9-11} \cmidrule(lr){12-14} \cmidrule(lr){15-17}
\textbf{Corpus} & \textbf{Method} & \textbf{B} & \textbf{S} & \textbf{BS} & \textbf{B} & \textbf{S} & \textbf{BS} & \textbf{B} & \textbf{S} & \textbf{BS} & \textbf{B} & \textbf{S} & \textbf{BS} & \textbf{B} & \textbf{S} & \textbf{BS} \\
\midrule
WikiLarge-FR & Direct & \textbf{34.52} & 36.25 & \textbf{0.65} & 33.86 & 36.48 & 0.61 & \textbf{32.25}$^{***}$ & 36.06 & \textbf{0.64}$^{***}$ & \textbf{33.49}$^{***}$ & 34.66 & \textbf{0.64}$^{***}$ & \textbf{34.66} & 36.16 & 0.65 \\
 & $T \rightarrow S$ C. & 23.03 & 36.07 & 0.58 & 19.80 & 35.51 & 0.53 & 23.88 & 36.42 & 0.60 & 19.16 & 35.09 & 0.56 & 31.42 & 36.80 & 0.64 \\
 & $S \rightarrow T$ C. & 31.98 & 35.46 & 0.62 & 32.75 & 36.85 & 0.61 & 28.21 & 37.17 & 0.61 & 25.72 & 35.03 & 0.59 & 32.49 & 36.03 & 0.65 \\
 & $T \rightarrow S$ D. & 19.77 & 35.68 & 0.57 & 23.83 & 37.25 & 0.57 & 25.20 & 37.00 & 0.60 & 17.94 & 34.38 & 0.54 & 19.15 & 35.16 & 0.56 \\
 & $S \rightarrow T$ D. & 24.28 & 36.30 & 0.59 & 26.78 & 37.78 & 0.58 & 25.64 & 36.57 & 0.60 & 29.64 & 34.73 & 0.61 & 29.17 & 35.96 & 0.62 \\
\midrule
CLEAR & Direct & 26.78 & 35.99 & \textbf{0.63} & 27.99 & 35.74 & 0.65 & 25.78 & 35.24 & 0.64 & \textbf{27.25}$^{**}$ & 34.36 & \textbf{0.64}$^{**}$ & 27.93 & 35.24 & 0.64 \\
 & $T \rightarrow S$ C. & 13.87 & 33.00 & 0.54 & 12.58 & 32.53 & 0.54 & 14.46 & 32.74 & 0.55 & 11.36 & 31.18 & 0.51 & 24.83 & 35.43 & 0.63 \\
 & $S \rightarrow T$ C. & 25.69 & 34.37 & 0.61 & 26.48 & 35.80 & 0.63 & 25.49 & 34.42 & 0.63 & 19.97 & 33.74 & 0.57 & 26.89 & 35.62 & 0.63 \\
 & $T \rightarrow S$ D. & 11.34 & 31.83 & 0.51 & 15.54 & 34.33 & 0.57 & 16.56 & 34.10 & 0.57 & 10.51 & 30.40 & 0.48 & 11.39 & 31.59 & 0.50 \\
 & $S \rightarrow T$ D. & 19.11 & 35.40 & 0.57 & 21.85 & 36.63 & 0.61 & 22.70 & 35.92 & 0.61 & 23.26 & 34.12 & 0.61 & 23.81 & 35.62 & 0.61 \\
\bottomrule
\end{tabular}%
}

\caption{Automatic Evaluation Metrics (\textbf{French to English}). Bold: significantly better than each other strategy ($p<0.05$, Welch's $t$-test); $^{**}$: $p<0.01$; $^{***}$: $p<0.001$. Abbreviations: B=BLEU, S=SARI, BS=BERTScore, C.=Composition, D.=Decomposition.}
\label{tab:french_metrics_compact}
\end{table*}

\section{Results}
\label{sec:results}

\subsection{Automatic Evaluation Metrics Results}
Performance of the different prompting approaches across datasets is provided in Table \ref{tab:english_metrics_compact} (English-to-French) and Table \ref{tab:french_metrics_compact} (French-to-English). As \texttt{Mistral-NeMo}, \texttt{Aya101}, and \texttt{GPT-OSS-20B} models generally yielded lower performance and are therefore less relevant to the strategy investigation, we present their results in Appendix \ref{app:additional_models} together with the \textbf{mT5} results (Appendix \ref{app:mt5}).

\paragraph{BLEU Score Performance:} Direct prompting achieves the highest BLEU score in 21 of the 25 corpus--model combinations across both translation directions, significantly outperforming the other approaches on the general-domain corpora. On ASSET, for instance, Direct reaches 66.1--70.1 BLEU with GPT-3.5, GPT-4o-mini and Gemini, compared with 47.1--60.5 for the four multi-step strategies with the same models. Translate-first approaches show the poorest BLEU performance in the French-to-English direction: on CLEAR, T$\rightarrow$S decomposition obtains only 10.5--16.6 BLEU across the five models, versus 25.8--28.0 for Direct, suggesting significant operation-ordering challenges for this language pair.

\paragraph{SARI Score Performance:} Multi-step approaches obtain the highest SARI scores across both directions and most models, with Simplify-then-Translate decomposition most often in the lead (e.g., 43.7 vs.\ 39.2 for Direct with GPT-3.5 on ASSET, and 38.9 vs.\ 32.3 on MultiCochrane). However, in almost all cases, the differences are not statistically significant.

\paragraph{Semantic Similarity Performance:} Analyses across both translation directions reveal distinct performance patterns and a notable directional asymmetry. For English-to-French tasks, direct prompting achieves significantly superior CamemBERT scores on general domain datasets (0.83--0.84 on ASSET vs.\ 0.74--0.83 for the multi-step strategies), suggesting strong semantic preservation for encyclopedic content. Conversely, in the French-to-English direction, while direct prompting remains superior, translate-first approaches yield the lowest BERTScores (on CLEAR, 0.48--0.57 for T$\rightarrow$S decomposition vs.\ 0.63--0.65 for Direct), indicating substantial semantic drift when French text is translated prior to simplification.

\paragraph{Sensitivity to MT-derived references:} For four corpora, the target-language references were obtained by machine translation, which may introduce translationese into the reference side~\cite{On_the_features_of_translationese}. Because all strategies are scored against the same references, this bias cannot by itself explain their relative ranking, and our feature-based and human results do not depend on references. As a check, we examine MultiCochrane, whose cross-lingual references are human-produced. The simplicity pattern is unchanged there: multi-step strategies obtain higher SARI than Direct, and a T$\rightarrow$S strategy obtains the highest Flesch Reading Ease, with all five main models. Direct's fidelity advantage, however, is smaller: it has the highest BLEU with three of the five models, and CamemBERTScore no longer separates the strategies consistently (Direct is highest with Gemini, lowest with GPT-3.5, and within 0.02 of the best strategy for the other three). 

The \textbf{mT5} baseline model achieved substantially lower BLEU and semantic scores across all datasets compared to the prompted LLMs (e.g., 9.7 BLEU on ASSET), though it showed competitive SARI performance on some corpora (37.5 on WikiAuto, above every prompted strategy).

\subsection{Human Evaluation Results}
Human annotations (full averages in Appendix~\ref{app:Human_annotation_res}) point to a trade-off. On the $-2$ to $2$ simplicity scale, Direct prompting receives the lowest ratings ($-0.06$ to $0.55$ across settings), while a decomposition strategy is rated simplest in six of the seven comparison settings, reaching 0.81 (WikiAuto, vs.\ source) and 0.83 (CLEAR, vs.\ translation) for T$\rightarrow$S decomposition. This increased simplicity is accompanied by more information removal: on the 1--5 scale, Direct stays between 1.02 and 1.20, whereas the multi-step strategies reach up to 1.74 (T$\rightarrow$S decomposition, MultiCochrane). Information addition remains low for all strategies (at most 1.28). Direct prompting thus removes the least information, indicating that it remains the most faithful to the source text.

These results should be read in light of inter-annotator agreement, computed as a quadratic weighted kappa with the simulation methodology of \citet{pavlick-tetreault-2016-empirical} (details in Appendix~\ref{app:Human_annotation_res}). Agreement is highest for information removal ($\kappa=0.360$) and limited for relative simplicity ($\kappa=0.216$) and information addition ($\kappa=0.192$). We therefore base our meaning-fidelity conclusions primarily on the removal dimension, and treat the simplicity and addition ratings as trends that corroborate, rather than independently establish, the linguistic-feature results of Section~\ref{sec:feature_results}. Lower agreement on simplicity than on meaning preservation has also been reported in monolingual simplification~\cite{sulem-etal-2018-simple}, and reflects the subjectivity of the notion, which depends on the reader's profile~\cite{grabar-saggion-2022-evaluation}; this subjectivity is precisely what motivates our complementary use of linguistic features as a more objective measure of simplicity.

\subsection{Feature-Based Analysis Results}
\label{sec:feature_results}

Our linguistic analysis, grounded in the feature set detailed in Table \ref{tab:linguistic_features}, revealed distinct patterns in how each prompting strategy achieves simplification, differentiating their impact across lexical, syntactic, and structural dimensions. Results of selected features appear in Appendix ~\ref{app:linguistic-features}.

The \textbf{Direct Prompting} approach consistently demonstrated the least significant changes in Syntactic and Structural Features (e.g., minimal reduction in 
syntactic tree depth and words per sentence). This approach produced outputs with higher complexity metrics, including elevated Flesch-Kincaid grade \citep{kincaid1975}, and lower Flesch Reading Ease \citep{flesch1948}, indicating higher complexity, and increased syntactic tree depth. Across the five main models and five corpora, Direct yields the lowest Flesch Reading Ease in 18 of 25 settings; for example, on MultiCochrane with GPT-4o-mini, Direct outputs average 29.7 words per sentence and a Flesch Reading Ease of 48.7, versus 22.1 and 62.6 for T$\rightarrow$S composition. These results indicate that direct prompting tends to generate more complex and formal language structures, suggesting a prioritization of fluency and translation fidelity over deep structural transformations. 

In contrast, the \textbf{composition} methods excelled at targeted complexity reduction by leveraging explicit intermediate steps. Specifically, 
\textbf{translate→simplify} generally resulted in simplified outputs, with higher Flesch Reading Ease scores (indicating better readability), reduced syntactic complexity, more sentences, higher short sentence ratios, and shorter sentence lengths. Overall, one of the two T$\rightarrow$S strategies (composition or decomposition) obtains the highest Flesch Reading Ease in 23 of the 25 model--corpus settings.
The \textbf{decomposition} approaches showed intermediate characteristics, with \textbf{simplify→translate decomposition} producing more complex outputs than \textbf{translate→simplify decomposition} (higher syntactic tree depth, words per sentence, and Flesch-Kincaid grade; lower Flesch Reading Ease in 23 of 25 settings, e.g., 47.2 vs.\ 58.9 with GPT-3.5 on CLEAR). This indicates that the order of operations in decomposition methods significantly influences the final output characteristics.

\subsection{Error Analysis}
\label{sec:error_analysis_main}

To reconcile the automatic, linguistic, and human evaluation findings, we annotated 500 GPT-3.5-Turbo outputs: 50 texts each from CLEAR (medical) and WikiLarge-FR (Wikipedia), under all five strategies. Each output was compared with the corpus reference simplification and labelled for four error types (Table~\ref{tab:error_counts}; examples in Appendix~\ref{app:error_analysis}).

\begin{table}[t]
\centering
\footnotesize
\setlength{\tabcolsep}{2pt}
{\renewcommand{\arraystretch}{1.05}%
\resizebox{\columnwidth}{!}{%
\begin{tabular}{@{}lccccc@{}}
\toprule
\textbf{Error Type} & \textbf{Direct} & \textbf{T\textrightarrow{}S C.} & \textbf{S\textrightarrow{}T C.} & \textbf{T\textrightarrow{}S D.} & \textbf{S\textrightarrow{}T D.} \\
\midrule
Information addition       & 14 & 7 & 10 & 4 & 6 \\
Failure to simplify        & 5  & 0 & 5  & 0 & 0 \\
Semantic inversion         & 4  & 1 & 1  & 0 & 0 \\
Propagated semantic error  & 0  & 0 & 0  & 2 & 2 \\
\midrule
\textbf{Total}             & \textbf{23} & \textbf{8} & \textbf{16} & \textbf{6} & \textbf{8} \\
\bottomrule
\end{tabular}%
}}
\caption{Error counts by strategy across 500 annotated outputs (50 per strategy per corpus on CLEAR and WikiLarge-FR, GPT-3.5-Turbo). \textbf{Abbreviations:} C.=Composition; D.=Decomposition.}
\label{tab:error_counts}
\end{table}

Direct prompting produces the most errors (23), driven mainly by unsupported \emph{information addition} (14) and \emph{failure to simplify} (5), i.e., outputs that are essentially translations, echoing \citet{joseph-etal-2023-multilingual}, who observed that direct multilingual simplification with GPT-3 yielded highly extractive outputs, in some cases word-for-word translations of the input. This source-preserving behaviour explains why Direct leads on BLEU and semantic similarity while showing the weakest readability gains. The T$\rightarrow$S strategies are the cleanest (6--8 errors) and never fail to simplify, consistent with their superior readability features. \emph{Semantic inversions} are rare (6 in total, 4 of them from Direct) but severe, since they leave BLEU and BERTScore largely unaffected; in the medical domain, for example, one Direct output replaced \emph{pneumonia} with \emph{urinary tract infection}. Finally, \emph{propagated semantic errors} (2 each) occur only in the decomposition strategies, where the second prompt has no access to the source text and cannot correct a flawed intermediate output. We further tested including the source text in the second prompt; a preliminary experiment with GPT-4o-mini (Appendix~\ref{app:source_aware}) suggests that this alone is insufficient: the source-aware variant restored some details dropped in the intermediate step and slightly raised BLEU, but it reproduced every propagated error of the standard pipeline (7 in both conditions), indicating that the model largely trusts the intermediate output.

\section{Conclusion}

Our evaluation of CLTS in English and French shows that no single prompting strategy outperforms the others on all evaluation dimensions; rather, the structure in which translation and simplification are combined determines a systematic trade-off between fidelity and simplicity. For applications prioritizing translation fidelity, Direct prompting is the most effective strategy, achieving the highest BLEU score in most settings across both language directions and all corpora and removing the least information on average according to human annotators. However, when accessibility and readability are paramount, translate-then-simplify approaches
demonstrate clear advantages, achieving the most effective syntactic simplification while maintaining moderate lexical richness and never failing to simplify in our error analysis.

The choice between single and multi-prompt execution also matters: decomposition strategies are uniquely exposed to propagated semantic errors, since the second prompt cannot consult the source text. These patterns recur across the five main LLMs, five corpora, and both translation directions, and persist under French-language and paraphrased instructions, indicating that they reflect the structure of the strategies rather than a particular model or wording. In practice, we recommend translate-then-simplify composition when readability is the priority, Direct prompting when fidelity is critical, and human review of outputs in high-stakes domains such as medicine.



\section*{Limitations}

While this study provides comprehensive insights into cross-lingual text simplification strategies, several limitations highlight directions for future work. First, our evaluation is restricted to a single task (CLTS) and a single, high-resource language pair (English--French), and our conclusions should be interpreted within the boundaries of this scenario. Whether the observed behaviours of the prompting strategies extend to other settings remains open; the primary direction for future work is therefore to apply our protocol to typologically more distant and lower-resource language pairs, as well as to other cross-lingual generation tasks that combine translation with a second transformation (e.g., summarization). Second, some nuanced aspects of simplification quality such as cultural appropriateness or actual reader comprehension are not captured by the automatic evaluation metrics (BLEU, SARI, BERTScore, and CamemBERTScore) and feature-based analysis used in the paper. Moreover, for four of the five corpora the target-language references were produced by machine translation; although the simplicity findings hold on the human-referenced MultiCochrane corpus (Section~\ref{sec:results}), absolute BLEU and semantic-similarity values on the other corpora may be affected by translationese in the references.  
Third, we study zero-shot instructions in order to isolate the structural variable (joint vs.\ sequential execution, operation order, and single- vs.\ multi-prompt packaging). Few-shot or chain-of-thought prompting, audience- or readability-conditioned prompts, fine-tuning, and iterative self-correction are not alternatives to these five structures but can be layered on top of any of them (e.g., few-shot T$\rightarrow$S composition); evaluating such combinations with our protocol is a natural follow-up.
Finally, while we evaluate nine LLMs spanning several model generations, the rapid evolution of these models means our findings represent a snapshot of current capabilities. Emerging models or fine-tuning approaches may yield different patterns and will thus require further evaluation and analysis, for which we provide an evaluation protocol in this paper.

\section*{Ethical Considerations}

CLTS aims to broaden access to information, but the same systems can mislead the users they intend to help. Our error analysis (Appendix~\ref{app:error_analysis}) identifies two risks of particular concern: rare but high-severity \emph{semantic errors} (meaning-reversing edits and propagated errors from decomposition pipelines), which remain undetected by BLEU and BERTScore, and frequent \emph{unsupported additions} that are indistinguishable from faithful simplification for readers unable to verify against the source. These risks are especially acute in medical content (CLEAR, MultiCochrane), where a fluent but incorrect output could directly cause harm. We therefore recommend human review in high-stakes domains, reporting SARI and feature-based readability alongside BLEU and semantic similarity, and clearly disclosing that outputs are machine-generated. 

All datasets were used in accordance with their original licenses; translations for constructing cross-lingual pairs were obtained via the Google Translate API for research use only; human annotation was conducted with informed in-house annotators, who were compensated for their work, and collected no personally identifying information.

\section*{Acknowledgements}
We thank the annotators for participating in our human evaluation experiment and the anonymous reviewers for their insightful comments. Our work was supported in part by grants by the Data Science Research Center and the Centre for the Study of Digital Politics and Strategy at Ben-Gurion University of the Negev, and by a BGU Grant for Interdisciplinary Research.

\bibliography{custom}


\clearpage
\appendix
\clearpage

\section{Model Documentation and Accessibility}
\label{sec:model_docs}

To ensure reproducibility and provide full transparency regarding the specific versions and technical documentation of the models used, we provide the official resource links for each LLM evaluated:
\begin{itemize}
    \item \textbf{GPT-3.5-Turbo}: \url{https://platform.openai.com/docs/models/gpt-3.5-turbo}
    \item \textbf{GPT-4o-Mini}: \url{https://platform.openai.com/docs/models/gpt-4o-mini}
    \item \textbf{GPT-5.4-Nano}: \url{https://developers.openai.com/api/docs/models/gpt-5.4-nano}
    \item \textbf{Gemini 2.5 Flash-Lite}: \url{https://deepmind.google/models/gemini/flash-lite/}
    \item \textbf{DeepSeek-Chat}: \url{https://api-docs.deepseek.com/news/news250821}
    \item \textbf{Mistral-NeMo}: \url{https://mistral.ai/news/mistral-nemo}
    \item \textbf{Aya-101}: \url{https://huggingface.co/CohereForAI/aya-101}
    \item \textbf{GPT-OSS-20B}: \url{https://openai.com/index/introducing-gpt-oss/}
\end{itemize}

To standardize the output format across all tested architectures, the models were prompted with a specific system instruction. No decoding parameters were set; all inferences used each provider's default decoding settings (e.g., temperature 1.0 and top-p 1.0 for the OpenAI and DeepSeek APIs).

To minimize conversational filler and ensure the models focused solely on the requested task, the following system prompt was utilized for all queries:
\begin{quote}
    \textit{``You are a text-to-text model. Your sole purpose is to provide the final output of a requested task. Do not include any interim steps, intermediate results, or conversational filler. Your response must begin directly with the final, complete answer.''}
\end{quote}

The technical specifications and specific snapshots for each model are detailed below:

\begin{itemize}
    \item \textbf{GPT-3.5 Turbo}: Version: \texttt{gpt-3.5-turbo-0125}; Parameters: Unknown; Context Window: 16,385 tokens; Max Output: 4,096 tokens; Knowledge Cutoff: Sept. 2021.
    \item \textbf{GPT-4o Mini}: Version: \texttt{gpt-4o-mini-2024-07-18}; Parameters: Unknown; Context Window: 128,000 tokens; Max Output: 16,384 tokens; Knowledge Cutoff: Oct. 2023.
    \item \textbf{GPT-5.4-Nano}: Version: \texttt{gpt-5.4-nano}; Parameters: Unknown; Context Window: 400,000 tokens; Max Output: 128,000 tokens; Knowledge Cutoff: Aug. 2025.
    \item \textbf{Gemini 2.5 Flash-Lite}: Version: \texttt{gemini-2.5-flash-lite} (Non-thinking); Parameters: Unknown; Context Window: 1,048,576 tokens; Max Output: 65,536 tokens; Knowledge Cutoff: Jan. 2025.
    \item \textbf{DeepSeek Chat}: Version: \texttt{DeepSeek-V3.1} (Non-thinking Mode); Parameters: 671B (MoE, 37B active); Context Window: 128,000 tokens; Max Output: 8,000 tokens; Knowledge Cutoff: March 2025.
    \item \textbf{Mistral NeMo}: Version: \texttt{open-mistral-nemo-2407}; Parameters: 12B; Context Window: 128,000 tokens; Max Output: Not reported; Knowledge Cutoff: Not reported.
    \item \textbf{Aya 101}: Version: \texttt{Aya-101}; Parameters: 13B (mT5-XXL architecture); Context Window: Not reported; Max Output: Not reported; Knowledge Cutoff: Not reported.
    \item \textbf{GPT-OSS-20B}: Version: \texttt{gpt-oss-20b}; Parameters: 21B (MoE, 3.6B active); Context Window: 128,000 tokens; Max Output: 131,072 tokens; Knowledge Cutoff: June 2024.
\end{itemize}

These details ensure that the environmental conditions of our experiments are fully transparent and replicable across different API snapshots and model iterations.

\section{mT5 Fine-tuning and Experimental Setup}
\label{sec:mt5_details}

To provide a robust baseline, we fine-tuned the \texttt{google/mt5-small} model. The specific hyperparameters used for the fine-tuning and inference of the mT5 model are detailed in Table~\ref{tab:mt5_hyperparams}.

\begin{table}[ht]
\centering
\small
\begin{tabular}{ll}
\hline
\textbf{Hyperparameter} & \textbf{Value} \\ \hline
Optimizer & AdamW \\
Training Epochs & 1 \\
Batch Size & 8 \\
Learning Rate & 0.001 \\
Gradient Accumulation & 4 steps \\
Max Input Length & 256 tokens \\
Max Output Length & 512 tokens \\
Decoding Strategy & Top-p Sampling \\
Top-p & 0.95 \\
Temperature & 1.0 \\ \hline
\end{tabular}
\caption{Fine-tuning and inference hyperparameters for the mT5 baseline.}
\label{tab:mt5_hyperparams}
\end{table}

\section{French-Prompt Robustness Experiment}
\label{app:french_prompt}

All prompting strategies in the main experiments are formulated in English (e.g., \textit{``Please simplify the following text in French:''}). To assess whether the language of the instruction itself influences results, we conducted a supplementary experiment in which all prompts were rewritten in French (e.g., \textit{``Veuillez simplifier le texte suivant en français:''}). This experiment used \textbf{GPT-4o-Mini} on the first 100 texts from each of the five corpora. The aggregated results are summarised in Table~\ref{tab:french_prompt_auto}, which reports the automatic evaluation metrics per corpus and prompting strategy, and Table~\ref{tab:french_prompt_features}, which reports the corresponding linguistic features.

\begin{table}[ht]
\centering
\footnotesize
\setlength{\tabcolsep}{4pt}
\resizebox{\columnwidth}{!}{%
\begin{tabular}{llccc}
\toprule
\textbf{Corpus} & \textbf{Strategy} & \textbf{BLEU} & \textbf{SARI} & \textbf{Sem.} \\
\midrule
\multirow{5}{*}{ASSET}
 & Direct            & 32.91 & 42.80 & 0.847 \\
 & T$\rightarrow$S Comp.    & 23.80 & 42.40 & 0.812 \\
 & S$\rightarrow$T Comp.    & 30.75 & 40.81 & 0.836 \\
 & T$\rightarrow$S Decomp.  & 27.31 & 42.98 & 0.832 \\
 & S$\rightarrow$T Decomp.  & 30.06 & 42.20 & 0.835 \\
\midrule
\multirow{5}{*}{WikiAuto}
 & Direct            & 28.16 & 33.75 & 0.782 \\
 & T$\rightarrow$S Comp.    & 20.78 & 34.81 & 0.755 \\
 & S$\rightarrow$T Comp.    & 26.47 & 33.57 & 0.782 \\
 & T$\rightarrow$S Decomp.  & 22.18 & 34.30 & 0.770 \\
 & S$\rightarrow$T Decomp.  & 26.93 & 34.99 & 0.775 \\
\midrule
\multirow{5}{*}{MultiCochrane}
 & Direct            & 12.29 & 34.39 & 0.658 \\
 & T$\rightarrow$S Comp.    & 8.82  & 39.23 & 0.629 \\
 & S$\rightarrow$T Comp.    & 11.53 & 35.17 & 0.654 \\
 & T$\rightarrow$S Decomp.  & 9.23  & 37.71 & 0.642 \\
 & S$\rightarrow$T Decomp.  & 11.17 & 36.08 & 0.655 \\
\midrule
\multirow{5}{*}{WikiLarge-FR}
 & Direct            & \textbf{44.20}$^{***}$ & 37.96 & \textbf{0.738}$^{**}$ \\
 & T$\rightarrow$S Comp.    & 28.42 & 37.21 & 0.664 \\
 & S$\rightarrow$T Comp.    & 28.86 & 35.95 & 0.663 \\
 & T$\rightarrow$S Decomp.  & 30.88 & 36.10 & 0.677 \\
 & S$\rightarrow$T Decomp.  & 30.13 & 35.81 & 0.669 \\
\midrule
\multirow{5}{*}{CLEAR}
 & Direct            & \textbf{26.97}$^{***}$ & 34.10 & \textbf{0.652}$^{**}$ \\
 & T$\rightarrow$S Comp.    & 13.37 & 32.83 & 0.547 \\
 & S$\rightarrow$T Comp.    & 17.83 & 34.24 & 0.590 \\
 & T$\rightarrow$S Decomp.  & 15.92 & 33.10 & 0.579 \\
 & S$\rightarrow$T Decomp.  & 18.28 & 34.35 & 0.586 \\
\bottomrule
\end{tabular}%
}
\caption{Automatic metrics (French-prompt experiment, GPT-4o-Mini, first 100 texts per corpus). Sem. = CamemBERTScore for EN$\rightarrow$FR corpora (ASSET, WikiAuto, MultiCochrane) and BERTScore for FR$\rightarrow$EN corpora (WikiLarge-FR, CLEAR). Bold: significantly better than each other strategy ($p<0.05$, Welch's $t$-test); $^{**}$: $p<0.01$; $^{***}$: $p<0.001$. \textbf{Abbreviations:} Comp.=Composition, Decomp.=Decomposition.}
\label{tab:french_prompt_auto}
\end{table}

\begin{table*}[ht]
\centering
\footnotesize
\setlength{\tabcolsep}{4pt}
\resizebox{\textwidth}{!}{%
\begin{tabular}{llccccccc}
\toprule
\textbf{Corpus} & \textbf{Prompt} & \textbf{Lex. Rich.} $\uparrow$ & \textbf{Tree Depth} $\downarrow$ & \textbf{Sent. \#} $\uparrow$ & \textbf{Words/Sent.} $\downarrow$ & \textbf{Short S.} $\uparrow$ & \textbf{FK Grade} $\downarrow$ & \textbf{FR Ease} $\uparrow$ \\
\midrule
\multirow{5}{*}{ASSET}
 & Direct            & 0.919 & 4.42 & 1.00 & 20.05 & 0.060 & 9.06 & 63.94 \\
 & T$\rightarrow$S Comp.    & 0.930 & 3.94 & 1.04 & 16.14 & 0.130 & \textbf{6.77} & \textbf{73.86} \\
 & S$\rightarrow$T Comp.    & 0.914 & 4.39 & 1.02 & 19.42 & 0.060 & 8.66 & 65.74 \\
 & T$\rightarrow$S Decomp.  & 0.925 & 4.20 & 1.01 & 17.90 & 0.120 & 7.92 & 68.61 \\
 & S$\rightarrow$T Decomp.  & 0.920 & 4.31 & 1.00 & 19.37 & 0.070 & 8.72 & 65.33 \\
\midrule
\multirow{5}{*}{WikiAuto}
 & Direct            & 0.918 & 4.30 & 1.03 & 19.75 & 0.090 & 8.88 & 65.18 \\
 & T$\rightarrow$S Comp.    & 0.930 & 3.85 & 1.04 & 16.48 & 0.125 & \textbf{6.99} & 73.24 \\
 & S$\rightarrow$T Comp.    & 0.914 & 4.30 & 1.04 & 19.71 & 0.100 & 8.78 & 65.97 \\
 & T$\rightarrow$S Decomp.  & 0.934 & 4.01 & 1.03 & 18.11 & 0.120 & 8.01 & 68.52 \\
 & S$\rightarrow$T Decomp.  & 0.915 & 4.14 & 1.02 & 19.07 & 0.080 & 8.69 & 65.56 \\
\midrule
\multirow{5}{*}{MultiCochrane}
 & Direct            & 0.884 & 5.34 & 1.03 & 25.93 & 0.055 & 12.48 & 49.57 \\
 & T$\rightarrow$S Comp.    & 0.911 & 4.78 & 1.04 & \textbf{19.98} & 0.085 & \textbf{9.21} & \textbf{62.99} \\
 & S$\rightarrow$T Comp.    & 0.888 & 5.26 & 1.06 & 25.17 & 0.065 & 11.77 & 53.41 \\
 & T$\rightarrow$S Decomp.  & 0.895 & 4.88 & 1.02 & 22.95 & 0.060 & 10.94 & 55.40 \\
 & S$\rightarrow$T Decomp.  & 0.893 & 5.27 & 1.04 & 24.50 & 0.055 & 11.81 & 51.90 \\
\midrule
\multirow{5}{*}{WikiLarge-FR}
 & Direct            & 0.919 & 5.63 & 1.00 & 18.67 & 0.080 & 9.84 & 57.13 \\
 & T$\rightarrow$S Comp.    & 0.944 & 5.01 & 1.01 & 15.49 & 0.130 & 7.53 & 68.24 \\
 & S$\rightarrow$T Comp.    & 0.937 & 5.08 & 1.01 & 15.43 & 0.145 & 7.71 & 66.78 \\
 & T$\rightarrow$S Decomp.  & 0.939 & 5.47 & 1.00 & 17.07 & 0.120 & 8.45 & 64.25 \\
 & S$\rightarrow$T Decomp.  & 0.924 & 5.25 & 1.01 & 16.38 & 0.145 & 8.46 & 63.15 \\
\midrule
\multirow{5}{*}{CLEAR}
 & Direct            & 0.950 & 5.99 & 1.00 & 18.21 & 0.140 & 12.05 & 40.06 \\
 & T$\rightarrow$S Comp.    & 0.971 & 4.97 & 1.02 & 14.05 & 0.240 & \textbf{9.41} & 52.18 \\
 & S$\rightarrow$T Comp.    & 0.952 & 5.25 & 1.00 & 15.84 & 0.200 & 10.57 & 47.13 \\
 & T$\rightarrow$S Decomp.  & 0.964 & 5.59 & 1.00 & 15.99 & 0.170 & 10.77 & 45.56 \\
 & S$\rightarrow$T Decomp.  & 0.947 & 5.51 & 1.02 & 15.99 & 0.175 & 11.10 & 42.98 \\
\bottomrule
\end{tabular}%
}
\caption{Linguistic features (French-prompt experiment, GPT-4o-Mini, first 100 texts per corpus). Lex. Rich. = lexical richness, Tree Depth = syntactic tree depth, Sent. \# = number of sentences, Words/Sent. = words per sentence, Short S. = short sentences ratio, FK Grade = Flesch-Kincaid Grade, FR Ease = Flesch Reading Ease. Arrows indicate the direction associated with simpler text. Bold: significantly simpler than each other strategy for the same corpus and feature ($p<0.05$, Welch's $t$-test). \textbf{Abbreviations:} Comp.=Composition, Decomp.=Decomposition.}
\label{tab:french_prompt_features}
\end{table*}

The results closely replicate the main findings. Direct prompting achieves the highest mean BLEU and semantic similarity on every corpus (tied with S$\rightarrow$T composition on WikiAuto semantic similarity); with 100 texts per corpus these advantages are statistically significant on the two FR$\rightarrow$EN corpora (WikiLarge-FR and CLEAR), but not on the EN$\rightarrow$FR corpora. Multi-step strategies obtain the highest SARI on four of five corpora, although none of the SARI differences is significant, and T$\rightarrow$S composition produces the most readable outputs, with the highest Flesch Reading Ease and lowest Flesch-Kincaid grade on every corpus and the lowest syntactic tree depth on every corpus. Every significant feature difference favours T$\rightarrow$S composition: its Flesch-Kincaid grade is significantly lower than that of every other strategy on four of the five corpora (all but WikiLarge-FR), its Flesch Reading Ease is significantly higher on ASSET and MultiCochrane, and its words per sentence significantly lower on MultiCochrane. This indicates that the key trends reported in Section~\ref{sec:results} are robust to the language in which the instructions are written.

\section{Prompt-Paraphrase Robustness Experiment}
\label{app:paraphrase}

Each strategy in the main experiments is realised by a single template, and all five templates share the frame ``Please \dots''. To test whether our conclusions depend on this particular phrasing, we re-ran all five strategies with \textbf{GPT-5.4-Nano} on the full test sets of all five corpora under two additional prompt sets. Within each set, all strategies share one frame, and the operation structure (order of operations and number of prompts) is identical to the original; only the wording changes. Both sets avoid the word ``simplify'', so a pattern that holds across sets cannot be attributed to a shared lexical cue. Prompts are shown for French as the target language; the same structure is used when English is the target.

\paragraph{V1 (bare imperative; terse, no politeness marker, lower information density).}
\begin{itemize}
    \item \textit{Direct:} ``Rewrite the following text in simple French:''
    \item \textit{T$\rightarrow$S Comp.:} ``Translate the following text into French, then rewrite your translation in simple French:''
    \item \textit{S$\rightarrow$T Comp.:} ``Rewrite the following text in simpler language, then translate your rewrite into French:''
    \item \textit{T$\rightarrow$S Decomp.:} Step 1: ``Translate the following text into French:''; Step 2: ``Rewrite the following text in simple French:''
    \item \textit{S$\rightarrow$T Decomp.:} Step 1: ``Rewrite the following text in simpler language:''; Step 2: ``Translate the following text into French:''
\end{itemize}

\paragraph{V2 (role and audience framing; higher information density).}
\begin{itemize}
    \item \textit{Direct:} ``You are an expert translator and text simplifier. Produce a French version of the text below that a reader without specialist knowledge can easily understand:''
    \item \textit{T$\rightarrow$S Comp.:} ``You are an expert translator and text simplifier. First produce a French translation of the text below; then rewrite that French translation so a reader without specialist knowledge can easily understand it:''
    \item \textit{S$\rightarrow$T Comp.:} ``You are an expert translator and text simplifier. First rewrite the text below so a reader without specialist knowledge can easily understand it; then translate that rewritten version into French:''
    \item \textit{T$\rightarrow$S Decomp.:} Step 1: ``You are an expert translator. Produce a French translation of the text below:''; Step 2: ``You are an expert text simplifier. Rewrite the French text below so a reader without specialist knowledge can easily understand it:''
    \item \textit{S$\rightarrow$T Decomp.:} Step 1: ``You are an expert text simplifier. Rewrite the text below so a reader without specialist knowledge can easily understand it:''; Step 2: ``You are an expert translator. Produce a French translation of the text below:''
\end{itemize}

Automatic metrics are reported in Table~\ref{tab:paraphrase_auto} and linguistic features in Tables~\ref{tab:paraphrase_feat_v1} and~\ref{tab:paraphrase_feat_v2}. Since the wording differs across sets, we compare strategies within each set.

\begin{table*}[ht]
\centering
\footnotesize
\setlength{\tabcolsep}{5pt}
\begin{tabular}{@{}llcccccc@{}}
\toprule
 & & \multicolumn{3}{c}{\textbf{V1 (bare imperative)}} & \multicolumn{3}{c}{\textbf{V2 (role/audience)}} \\
\cmidrule(lr){3-5} \cmidrule(lr){6-8}
\textbf{Corpus} & \textbf{Strategy} & \textbf{BLEU} & \textbf{SARI} & \textbf{Sem.} & \textbf{BLEU} & \textbf{SARI} & \textbf{Sem.} \\
\midrule
\multirow{5}{*}{ASSET}
 & Direct                  & 25.1 & 37.8 & 0.823 & \textbf{22.7}$^{**}$ & 38.0 & \textbf{0.813}$^{**}$ \\
 & T$\rightarrow$S Comp.   & 21.3 & 39.8 & 0.801 & 16.8 & 39.0 & 0.774 \\
 & S$\rightarrow$T Comp.   & 23.5 & 40.4 & 0.813 & 18.7 & 38.6 & 0.786 \\
 & T$\rightarrow$S Decomp. & 22.5 & 39.4 & 0.812 & 16.9 & 37.2 & 0.758 \\
 & S$\rightarrow$T Decomp. & 23.9 & 40.6 & 0.814 & 19.1 & 39.2 & 0.788 \\
\midrule
\multirow{5}{*}{WikiAuto}
 & Direct                  & \textbf{26.4}$^{***}$ & 34.3 & \textbf{0.790}$^{***}$ & \textbf{23.9}$^{***}$ & 34.3 & \textbf{0.781}$^{***}$ \\
 & T$\rightarrow$S Comp.   & 20.2 & 34.3 & 0.765 & 15.8 & 33.4 & 0.748 \\
 & S$\rightarrow$T Comp.   & 22.0 & 34.8 & 0.771 & 18.9 & 34.4 & 0.759 \\
 & T$\rightarrow$S Decomp. & 23.0 & 34.5 & 0.778 & 17.6 & 33.5 & 0.743 \\
 & S$\rightarrow$T Decomp. & 23.0 & 35.1 & 0.773 & 18.5 & 34.4 & 0.753 \\
\midrule
\multirow{5}{*}{MultiCochrane}
 & Direct                  & 10.0 & 36.9 & 0.659 & \textbf{8.3}$^{**}$ & 38.1 & 0.649 \\
 & T$\rightarrow$S Comp.   & 6.9 & 38.2 & 0.607 & 5.4 & 38.4 & 0.606 \\
 & S$\rightarrow$T Comp.   & 8.5 & 39.2 & 0.646 & 6.6 & 39.0 & 0.630 \\
 & T$\rightarrow$S Decomp. & 8.5 & 38.3 & 0.641 & 5.5 & 37.4 & 0.600 \\
 & S$\rightarrow$T Decomp. & 8.6 & 39.2 & 0.637 & 6.6 & 39.6 & 0.619 \\
\midrule
\multirow{5}{*}{CLEAR}
 & Direct                  & 19.5 & 34.8 & 0.590 & \textbf{19.2}$^{**}$ & 34.5 & \textbf{0.582}$^{***}$ \\
 & T$\rightarrow$S Comp.   & 11.4 & 31.6 & 0.510 & 8.1 & 30.2 & 0.451 \\
 & S$\rightarrow$T Comp.   & 19.0 & 35.0 & 0.583 & 15.8 & 34.4 & 0.541 \\
 & T$\rightarrow$S Decomp. & 14.2 & 33.3 & 0.544 & 10.3 & 31.9 & 0.484 \\
 & S$\rightarrow$T Decomp. & 18.6 & 35.0 & 0.577 & 13.5 & 33.8 & 0.516 \\
\midrule
\multirow{5}{*}{WikiLarge-FR}
 & Direct                  & \textbf{28.5}$^{***}$ & 36.8 & \textbf{0.619}$^{**}$ & \textbf{28.2}$^{***}$ & 36.6 & \textbf{0.617}$^{***}$ \\
 & T$\rightarrow$S Comp.   & 20.9 & 36.3 & 0.576 & 16.3 & 34.8 & 0.533 \\
 & S$\rightarrow$T Comp.   & 24.3 & 36.7 & 0.589 & 21.6 & 35.8 & 0.567 \\
 & T$\rightarrow$S Decomp. & 23.8 & 37.2 & 0.592 & 18.7 & 35.9 & 0.552 \\
 & S$\rightarrow$T Decomp. & 24.9 & 36.9 & 0.593 & 20.0 & 36.0 & 0.552 \\
\bottomrule
\end{tabular}
\caption{Automatic metrics for the two paraphrased prompt sets (GPT-5.4-Nano, full test sets). Bold: significantly better than each other strategy for the same corpus and metric ($p<0.05$, Welch's $t$-test); $^{**}$: $p<0.01$; $^{***}$: $p<0.001$. Sem. = CamemBERTScore for EN$\rightarrow$FR corpora (ASSET, WikiAuto, MultiCochrane) and BERTScore for FR$\rightarrow$EN corpora (CLEAR, WikiLarge-FR). \textbf{Abbreviations:} Comp.=Composition, Decomp.=Decomposition.}
\label{tab:paraphrase_auto}
\end{table*}

\begin{table*}[ht]
\centering
\footnotesize
\setlength{\tabcolsep}{2pt}
\begin{tabular}{@{}llccccccc@{}}
\toprule
\textbf{Corpus} & \textbf{Strategy} & \textbf{Lex. Rich.} $\uparrow$ & \textbf{Tree Depth} $\downarrow$ & \textbf{Sent. \#} $\uparrow$ & \textbf{Words/Sent.} $\downarrow$ & \textbf{Short S.} $\uparrow$ & \textbf{FK Grade} $\downarrow$ & \textbf{FR Ease} $\uparrow$ \\
\midrule
\multirow{5}{*}{ASSET}
 & Direct                  & 0.90 & 4.45 & 1.18 & 20.9 & 0.041 & 9.5 & 61.7 \\
 & T$\rightarrow$S Comp.   & 0.90 & 4.13 & \textbf{1.52} & \textbf{16.3} & \textbf{0.136} & \textbf{7.1} & \textbf{71.3} \\
 & S$\rightarrow$T Comp.   & 0.90 & 4.40 & 1.28 & 19.1 & 0.063 & 8.8 & 64.4 \\
 & T$\rightarrow$S Decomp. & 0.90 & 4.31 & 1.31 & 19.4 & 0.067 & 8.6 & 65.4 \\
 & S$\rightarrow$T Decomp. & 0.90 & 4.43 & 1.25 & 19.7 & 0.076 & 9.1 & 62.5 \\
\midrule
\multirow{5}{*}{WikiAuto}
 & Direct                  & 0.89 & 4.31 & 1.31 & 20.1 & 0.068 & 9.1 & 63.5 \\
 & T$\rightarrow$S Comp.   & 0.89 & \textbf{3.97} & \textbf{1.69} & \textbf{15.4} & \textbf{0.154} & \textbf{6.8} & \textbf{72.3} \\
 & S$\rightarrow$T Comp.   & 0.89 & 4.26 & 1.46 & 18.3 & 0.099 & 8.4 & 66.0 \\
 & T$\rightarrow$S Decomp. & 0.89 & 4.24 & 1.45 & 18.6 & 0.082 & 8.4 & 66.3 \\
 & S$\rightarrow$T Decomp. & 0.89 & 4.32 & 1.40 & 18.8 & 0.078 & 8.7 & 64.3 \\
\midrule
\multirow{5}{*}{MultiCochrane}
 & Direct                  & 0.87 & 5.55 & 1.25 & 26.2 & 0.048 & 12.0 & 53.3 \\
 & T$\rightarrow$S Comp.   & 0.88 & 5.09 & \textbf{1.54} & \textbf{21.3} & 0.086 & \textbf{8.7} & \textbf{69.1} \\
 & S$\rightarrow$T Comp.   & 0.87 & 5.24 & 1.41 & 23.5 & 0.059 & 10.6 & 58.5 \\
 & T$\rightarrow$S Decomp. & 0.87 & 5.56 & 1.40 & 24.6 & 0.041 & 11.0 & 58.0 \\
 & S$\rightarrow$T Decomp. & 0.88 & 5.40 & 1.34 & 24.6 & 0.038 & 11.4 & 54.8 \\
\midrule
\multirow{5}{*}{CLEAR}
 & Direct                  & 0.93 & 5.90 & 1.16 & 18.2 & 0.108 & 11.3 & 44.9 \\
 & T$\rightarrow$S Comp.   & 0.93 & 5.46 & 1.27 & 16.8 & \textbf{0.165} & \textbf{9.2} & \textbf{58.2} \\
 & S$\rightarrow$T Comp.   & 0.92 & 5.94 & 1.21 & 17.9 & 0.118 & 11.3 & 44.5 \\
 & T$\rightarrow$S Decomp. & 0.93 & 5.75 & 1.20 & 17.6 & 0.111 & 10.1 & 52.3 \\
 & S$\rightarrow$T Decomp. & 0.93 & 6.07 & 1.16 & 18.3 & 0.100 & 11.7 & 42.6 \\
\midrule
\multirow{5}{*}{WikiLarge-FR}
 & Direct                  & 0.88 & 5.34 & 1.39 & 16.6 & 0.121 & 9.0 & 59.2 \\
 & T$\rightarrow$S Comp.   & 0.89 & \textbf{5.04} & \textbf{1.57} & \textbf{14.4} & \textbf{0.187} & \textbf{7.5} & \textbf{66.1} \\
 & S$\rightarrow$T Comp.   & 0.89 & 5.29 & 1.42 & 16.2 & 0.139 & 8.8 & 60.0 \\
 & T$\rightarrow$S Decomp. & 0.89 & 5.23 & 1.46 & 15.7 & 0.129 & 8.3 & 62.6 \\
 & S$\rightarrow$T Decomp. & 0.89 & 5.47 & 1.35 & 17.2 & 0.104 & 9.5 & 56.6 \\
\bottomrule
\end{tabular}
\caption{Linguistic features for prompt set \textbf{V1} (bare imperative; GPT-5.4-Nano). Bold: significantly simpler than each other strategy for the same corpus and feature ($p<0.05$, Welch's $t$-test). Arrows indicate the direction associated with simpler text. \textbf{Abbreviations:} as in Table~\ref{tab:french_prompt_features}.}
\label{tab:paraphrase_feat_v1}
\end{table*}

\begin{table*}[ht]
\centering
\footnotesize
\setlength{\tabcolsep}{2pt}
\begin{tabular}{@{}llccccccc@{}}
\toprule
\textbf{Corpus} & \textbf{Strategy} & \textbf{Lex. Rich.} $\uparrow$ & \textbf{Tree Depth} $\downarrow$ & \textbf{Sent. \#} $\uparrow$ & \textbf{Words/Sent.} $\downarrow$ & \textbf{Short S.} $\uparrow$ & \textbf{FK Grade} $\downarrow$ & \textbf{FR Ease} $\uparrow$ \\
\midrule
\multirow{5}{*}{ASSET}
 & Direct                  & 0.90 & 4.57 & 1.24 & 21.3 & 0.035 & 9.7 & 60.8 \\
 & T$\rightarrow$S Comp.   & 0.89 & 4.57 & 1.40 & 20.0 & 0.048 & 8.6 & 66.6 \\
 & S$\rightarrow$T Comp.   & 0.88 & 4.64 & 1.44 & 20.5 & 0.043 & 9.3 & 62.3 \\
 & T$\rightarrow$S Decomp. & 0.89 & 4.56 & 1.43 & 19.4 & 0.055 & 8.8 & 64.5 \\
 & S$\rightarrow$T Decomp. & 0.89 & 4.59 & 1.36 & 20.9 & 0.039 & 9.5 & 61.5 \\
\midrule
\multirow{5}{*}{WikiAuto}
 & Direct                  & 0.89 & 4.45 & 1.39 & 20.2 & 0.063 & 9.2 & 63.1 \\
 & T$\rightarrow$S Comp.   & 0.89 & 4.38 & 1.60 & 18.5 & 0.076 & \textbf{8.1} & \textbf{68.1} \\
 & S$\rightarrow$T Comp.   & 0.88 & 4.52 & 1.60 & 19.2 & 0.080 & 8.8 & 64.4 \\
 & T$\rightarrow$S Decomp. & 0.89 & 4.35 & 1.60 & 18.5 & 0.085 & 8.4 & 66.2 \\
 & S$\rightarrow$T Decomp. & 0.88 & 4.55 & 1.58 & 19.1 & 0.075 & 8.8 & 63.9 \\
\midrule
\multirow{5}{*}{MultiCochrane}
 & Direct                  & 0.86 & 5.59 & 1.38 & 26.5 & 0.032 & 12.1 & 53.4 \\
 & T$\rightarrow$S Comp.   & 0.87 & 5.66 & 1.60 & 23.8 & 0.047 & 10.1 & \textbf{62.9} \\
 & S$\rightarrow$T Comp.   & 0.86 & 5.71 & 1.64 & 24.7 & 0.050 & 11.2 & 56.5 \\
 & T$\rightarrow$S Decomp. & 0.86 & 5.62 & 1.63 & 23.7 & 0.044 & 10.7 & 59.2 \\
 & S$\rightarrow$T Decomp. & 0.86 & 5.89 & 1.57 & 25.7 & 0.044 & 11.8 & 54.0 \\
\midrule
\multirow{5}{*}{CLEAR}
 & Direct                  & 0.92 & 6.26 & 1.12 & 19.7 & 0.097 & 12.2 & 41.5 \\
 & T$\rightarrow$S Comp.   & 0.92 & 6.22 & 1.28 & 19.9 & 0.079 & 10.3 & 55.4 \\
 & S$\rightarrow$T Comp.   & 0.91 & 6.28 & 1.28 & 19.2 & 0.083 & 11.5 & 45.4 \\
 & T$\rightarrow$S Decomp. & 0.92 & 6.02 & 1.28 & 18.8 & 0.061 & 10.3 & 53.9 \\
 & S$\rightarrow$T Decomp. & 0.91 & 6.42 & 1.28 & 19.7 & 0.097 & 11.7 & 44.7 \\
\midrule
\multirow{5}{*}{WikiLarge-FR}
 & Direct                  & 0.89 & 5.71 & 1.28 & 18.4 & 0.079 & 10.0 & 54.4 \\
 & T$\rightarrow$S Comp.   & 0.89 & 5.62 & 1.48 & 17.3 & 0.090 & 8.6 & 62.8 \\
 & S$\rightarrow$T Comp.   & 0.88 & 5.62 & 1.49 & 17.4 & 0.101 & 9.4 & 57.5 \\
 & T$\rightarrow$S Decomp. & 0.88 & 5.56 & 1.53 & \textbf{16.7} & 0.112 & 8.6 & 61.7 \\
 & S$\rightarrow$T Decomp. & 0.88 & 5.71 & 1.50 & 17.6 & 0.095 & 9.5 & 57.1 \\
\bottomrule
\end{tabular}
\caption{Linguistic features for prompt set \textbf{V2} (role/audience framing; GPT-5.4-Nano). Bold: significantly simpler than each other strategy for the same corpus and feature ($p<0.05$, Welch's $t$-test). Arrows indicate the direction associated with simpler text. \textbf{Abbreviations:} as in Table~\ref{tab:french_prompt_features}.}
\label{tab:paraphrase_feat_v2}
\end{table*}

\paragraph{Findings.} The source-fidelity conclusion is preserved: Direct prompting obtains the highest BLEU and the highest semantic similarity on every corpus under both sets, and this advantage is significant in 13 of the 20 comparisons: under V2 on every corpus for BLEU and on all but MultiCochrane for semantic similarity, and under V1 on WikiAuto and WikiLarge-FR. The simplification conclusion is also preserved. T$\rightarrow$S composition obtains the highest Flesch Reading Ease on every corpus under both sets, significantly so in 7 of the 10 corpus--set combinations (all five corpora under V1; WikiAuto and MultiCochrane under V2), and Direct obtains the lowest Flesch Reading Ease in 8 of the 10 combinations. Under V1, T$\rightarrow$S composition is significantly simpler than every other strategy on 24 of the 30 corpus--feature cells for tree depth, sentence count, words per sentence, short-sentence ratio, Flesch-Kincaid grade, and Flesch Reading Ease. Under V2, differences on the individual features are smaller and mostly not significant; the few significant effects all favour a T$\rightarrow$S strategy (Flesch Reading Ease on WikiAuto and MultiCochrane and Flesch-Kincaid grade on WikiAuto for T$\rightarrow$S composition; words per sentence on WikiLarge-FR for T$\rightarrow$S decomposition). As in the main experiments, SARI differences between strategies are small and never significant. The core trade-off, in which Direct maximises fidelity and translate-then-simplify maximises readability, therefore holds across both paraphrased prompt sets.

\section{Human Annotation Guidelines}
\label{app:annotation_guidelines}
Annotators were presented with a side-by-side view of the original text or its translated version (Version A) and the simplified LLM response (Version B). They were provided with the following prompt: 
\textit{``Given Versions A and B, please answer the following questions:''}

Following prior work on simplification evaluation~\cite{nisioi-etal-2017-exploring,sulem-etal-2018-simple}, each question is phrased as a yes/no question, and annotators answer it on a discrete scale. For simplicity, each point of the scale is tied to an explicit answer; for information addition and removal, the endpoints of the 1--5 scale are anchored and intermediate values express increasing intensity.

\begin{itemize}
    \item \textbf{Simplicity:} \textit{``Is B simpler than A?''}
    \\ Scale: $+2$ = B is much simpler than A; $+1$ = B is quite simpler than A; $0$ = A and B are equally complex; $-1$ = B is quite more complex than A; $-2$ = B is much more complex than A.
    
    \item \textbf{Information Addition:} \textit{``Does Version B add information compared to Version A?''} 
    \\ (Scale: 1 to 5, where 1 is no addition and 5 is significant addition).

    \item \textbf{Information Removal:} \textit{``Does Version B remove information compared to Version A?''} 
    \\ (Scale: 1 to 5, where 1 is no removal and 5 is total loss of original meaning).

\end{itemize}

\section{Linguistic Features Descriptions}
\label{features_descriptions}

This appendix provides a detailed glossary of the linguistic features used in our feature-based analysis, organized by category (lexical, syntactic, readability, named entity, and grammatical), with a short description of how each feature is computed. The full glossary is presented in Table~\ref{tab:linguistic_features_glossary}.

\begin{table*}[!ht]
\small
\centering
\begin{tabularx}{\textwidth}{l l X}
\toprule
\textbf{Category} & \textbf{Feature} & \textbf{Description} \\
\midrule
\multirow{5}{*}{\textbf{Lexical}} & Lexical Richness & Unique types divided by total words. \\
 & Infrequent Words Ratio & Ratio of words not in a language-specific high-frequency reference list. \\
 & Long Words Ratio & Words with $\ge$ 8 characters per total words. \\
 & Content Words Ratio & Nouns/verbs/adj/adv per total words. \\
 & Average Word Length & Mean characters per word. \\
\midrule
\multirow{11}{*}{\textbf{Syntactic}} & Words before Main Verb & Tokens preceding sentence head. \\
 & Noun Phrases Ratio & Number of noun phrases per total tokens. \\
 & Relative Clauses Ratio & Number of relative clauses per total tokens. \\
 & Appositions Ratio & Number of appositive constructions per total tokens. \\
 & Conditional Clauses Ratio & Number of conditional clauses per total tokens. \\
 & Conjunctions Ratio & Number of coordinating/subordinating conjunctions per total tokens. \\
 & Passive Voice Ratio & Number of passive-voice constructions per total verbs. \\
 & Syntactic Tree Depth & Max dependency tree depth. \\
 & Sentences Number & Total sentence count. \\
 & Words per Sentence & Mean words per sentence. \\
 & Short Sentences Ratio & Ratio of sentences under 10 words. \\
\midrule
\multirow{3}{*}{\textbf{Readability}} & Flesch Reading Ease & Score calculating ease of reading (higher = easier). \\
 & Flesch Kincaid Grade & U.S. school grade level required to read the text. \\
 & Syllables Ratio & Total syllables divided by total words. \\
\midrule
\multirow{7}{*}{\textbf{Named Entity}}
 & Max Same Entity Distance & Maximum token distance between two mentions of the same entity. \\
 & Unique Entities & Total distinct entity count. \\
 & Unique Entities Average & Mean number of unique entities per sentence. \\
 & Avg Same Entity Distance & Mean tokens between mentions of the same entity. \\
 & Entity to Token Ratio & Entity density per total tokens. \\
 & Unique Entities to Total Entities & Distinct entities per total entities. \\
 & Consecutive Entity Distance & Mean token distance between consecutive entity mentions. \\
\midrule
\multirow{6}{*}{\textbf{Grammatical}}
 & Modifiers Ratio & Number of modifiers (adjectives/adverbs) per total tokens. \\
 & Negations Ratio & Number of negation markers per total tokens. \\
 & Past Tense Verbs Ratio & Total past tense verbs divided by total number of verbs. \\
 & Punctuation Ratio & Punctuation marks per total tokens. \\
 & Third Person Pronouns Ratio & Number of third-person pronouns per total tokens. \\
\bottomrule
\end{tabularx}
\caption{Detailed glossary of the linguistic features categorized by type.}
\label{tab:linguistic_features_glossary}
\end{table*}

\section{Open Source Models Results}
\label{app:additional_models}

This appendix reports the automatic evaluation metrics (BLEU, SARI, and semantic similarity) for the three additional open-source models, Aya101, Mistral-NeMo, and GPT-OSS-20B, across all five corpora and all five prompting strategies, complementing the main-text results restricted to the five primary models. The full results are reported in Table~\ref{app:tab:additional_models_metrics_all}.

\begin{table*}[ht]
\centering
\footnotesize
\setlength{\tabcolsep}{2pt} 
\begin{tabular}{@{}llccccccccc@{}}
\toprule
 & & \multicolumn{3}{c}{\textbf{Aya101}} & \multicolumn{3}{c}{\textbf{Mistral-NeMo}} & \multicolumn{3}{c}{\textbf{GPT-OSS-20B}} \\
\cmidrule(lr){3-5} \cmidrule(lr){6-8} \cmidrule(lr){9-11}
\textbf{Corpus} & \textbf{Method} & \textbf{BLEU} & \textbf{SARI} & \textbf{Sem.} & \textbf{BLEU} & \textbf{SARI} & \textbf{Sem.} & \textbf{BLEU} & \textbf{SARI} & \textbf{Sem.} \\
\midrule
\multirow{5}{*}{ASSET}
 & Direct & 54.12 & 39.66 & 0.79 & 56.40 & 38.58 & 0.78 & 59.11 & 42.12 & 0.81 \\
 & $T \rightarrow S$ Composition & 56.22 & 38.47 & 0.80 & 65.88 & 38.94 & 0.82 & 52.61 & 41.97 & 0.78 \\
 & $S \rightarrow T$ Composition & 56.37 & 38.83 & 0.80 & 64.82 & 37.91 & 0.83 & 57.12 & 43.30 & 0.80 \\
 & $T \rightarrow S$ Decomposition & 55.91 & 39.07 & 0.79 & 54.21 & \textbf{40.69} & 0.77 & 54.47 & 43.16 & 0.80 \\
 & $S \rightarrow T$ Decomposition & 54.20 & 38.46 & 0.76 & 46.92 & 39.56 & 0.73 & 57.75 & 43.93 & 0.80 \\
\midrule
\multirow{5}{*}{WikiAuto}
 & Direct & 25.10 & 32.88 & 0.76 & 23.63 & 31.50 & 0.74 & \textbf{25.09}$^{***}$ & 34.32 & \textbf{0.78}$^{***}$ \\
 & $T \rightarrow S$ Composition & 26.91 & 32.42 & 0.76 & 28.74 & 32.08 & 0.78 & 20.51 & 33.95 & 0.75 \\
 & $S \rightarrow T$ Composition & 26.61 & 32.56 & 0.76 & \textbf{30.58}$^{***}$ & 31.93 & 0.78 & 22.05 & 34.32 & 0.76 \\
 & $T \rightarrow S$ Decomposition & 25.65 & 32.50 & 0.76 & 21.07 & \textbf{32.81}$^{**}$ & 0.73 & 22.02 & 34.30 & 0.77 \\
 & $S \rightarrow T$ Decomposition & 24.29 & 33.29 & 0.73 & 16.22 & 31.93 & 0.70 & 22.41 & 34.43 & 0.76 \\
\midrule
\multirow{5}{*}{MultiCochrane}
 & Direct & 9.58 & 34.56 & 0.63 & 10.44 & 34.20 & 0.62 & 9.56 & 37.14 & 0.64 \\
 & $T \rightarrow S$ Composition & 10.13 & 33.02 & 0.63 & 12.86 & 32.82 & 0.66 & 8.04 & 38.44 & 0.61 \\
 & $S \rightarrow T$ Composition & 9.88 & 33.24 & 0.63 & 12.93 & 31.32 & 0.65 & 8.17 & 38.60 & 0.63 \\
 & $T \rightarrow S$ Decomposition & 9.66 & 33.31 & 0.63 & 9.66 & 37.24 & 0.62 & 8.97 & 38.61 & 0.64 \\
 & $S \rightarrow T$ Decomposition & 9.48 & 33.84 & 0.62 & 6.11 & 37.13 & 0.56 & 8.41 & 39.09 & 0.63 \\
\midrule
\multirow{5}{*}{WikiLarge-FR}
 & Direct & 31.98 & 34.33 & 0.63 & 28.09 & 34.69 & 0.61 & \textbf{28.62}$^{***}$ & 36.35 & \textbf{0.63} \\
 & $T \rightarrow S$ Composition & 33.92 & 33.74 & 0.64 & 31.62 & 35.62 & 0.63 & 21.22 & 35.62 & 0.57 \\
 & $S \rightarrow T$ Composition & 33.20 & 33.86 & 0.64 & \textbf{35.37}$^{***}$ & 34.64 & \textbf{0.65} & 25.37 & 36.23 & 0.60 \\
 & $T \rightarrow S$ Decomposition & 32.70 & 33.84 & 0.63 & 16.91 & 32.98 & 0.52 & 23.54 & 36.74 & 0.60 \\
 & $S \rightarrow T$ Decomposition & 29.48 & 34.73 & 0.60 & 26.92 & 34.79 & 0.60 & 25.70 & 36.17 & 0.60 \\
\midrule
\multirow{5}{*}{CLEAR}
 & Direct & 25.57 & 35.16 & 0.59 & 20.94 & 34.39 & 0.60 & 19.84 & 33.81 & 0.60 \\
 & $T \rightarrow S$ Composition & 26.27 & 33.82 & 0.60 & 23.34 & 35.00 & 0.61 & 12.16 & 31.59 & 0.53 \\
 & $S \rightarrow T$ Composition & 26.44 & 34.84 & 0.61 & \textbf{28.22}$^{***}$ & 34.15 & \textbf{0.64} & 19.13 & 35.04 & 0.58 \\
 & $T \rightarrow S$ Decomposition & 26.48 & 34.80 & 0.60 & 10.18 & 30.25 & 0.46 & 13.05 & 31.89 & 0.54 \\
 & $S \rightarrow T$ Decomposition & 23.15 & 33.92 & 0.55 & 20.30 & 34.28 & 0.58 & 17.74 & 34.72 & 0.57 \\
\bottomrule
\end{tabular}
\caption{Automatic Evaluation Metrics for additional models Aya101, Mistral-NeMo, and GPT-OSS-20B across all corpora. Bold: significantly better than each other strategy ($p<0.05$, Welch's $t$-test); $^{**}$: $p<0.01$; $^{***}$: $p<0.001$. \textbf{Abbreviations:} Sem.: semantic similarity metrics (BERTScore or CamemBERTScore).}
\label{app:tab:additional_models_metrics_all}
\end{table*}

\section{mT5 Results}
\label{app:mt5}

This appendix reports the BLEU, SARI, and semantic-similarity scores obtained by our fine-tuned mT5 baseline across all five corpora, serving as a reference point against which the prompted LLM strategies in the main text are compared. As mT5 is fine-tuned end-to-end on the cross-lingual simplification task and does not consume the five prompting strategies, we report a single score per corpus rather than a per-strategy breakdown. The per-corpus scores are presented in Table~\ref{tab:mt5_results}.

\begin{table}[ht]
\centering
\small
\begin{tabular}{lccc}
\toprule
\textbf{Corpus} & \textbf{BLEU} & \textbf{SARI} & \textbf{Sem.} \\
\midrule
ASSET           & 9.678  & 28.817 & 0.331  \\
WikiAuto        & 11.213 & 37.525 & 0.496  \\
MultiCochrane   & 2.529  & 35.436 & 0.269  \\
CLEAR           & 3.214  & 24.708 & -0.004 \\
WikiLarge-FR    & 8.700  & 29.658 & 0.132  \\
\bottomrule
\end{tabular}
\caption{mT5 model evaluation results across different corpora and metrics. Sem.: Semantic Score, computed as CamemBERTScore for English$\rightarrow$French simplification and BERTScore for French$\rightarrow$English simplification.}
\label{tab:mt5_results}
\end{table}

\section{Human Annotation Results}
\label{app:Human_annotation_res}

This appendix presents the per-corpus average human ratings for each prompting strategy along the three annotation dimensions, relative simplicity, added information scale, and removed information scale. The averaged ratings are reported in Table~\ref{tab:human_eval_small}.

\begin{table}[!ht]
\centering
\footnotesize
\setlength{\tabcolsep}{3.5pt} 
\begin{tabular}{@{}llccc@{}}
\toprule
\textbf{Corpus} & \textbf{Strategy} & \textbf{Simp.} & \textbf{Add.} & \textbf{Rem.} \\ \midrule
\multirow{5}{*}{ASSET} & Direct & -0.03 & 1.06 & 1.08 \\
 & $T \rightarrow S$ Comp. & 0.04 & 1.09 & 1.56 \\
 & $S \rightarrow T$ Comp. & 0.09 & 1.08 & 1.55 \\
 & $T \rightarrow S$ Decomp. & 0.07 & 1.23 & 1.27 \\
 & $S \rightarrow T$ Decomp. & 0.14 & 1.26 & 1.37 \\ \cmidrule(l){1-5}
\multirow{5}{*}{M-Cochrane} & Direct & 0.06 & 1.01 & 1.20 \\
 & $T \rightarrow S$ Comp. & 0.16 & 1.00 & 1.14 \\
 & $S \rightarrow T$ Comp. & 0.19 & 1.01 & 1.15 \\
 & $T \rightarrow S$ Decomp. & 0.51 & 1.01 & 1.74 \\
 & $S \rightarrow T$ Decomp. & 0.64 & 1.01 & 1.55 \\ \cmidrule(l){1-5}
\multirow{5}{*}{WikiLarge-FR} & Direct & 0.15 & 1.02 & 1.03 \\
 & $T \rightarrow S$ Comp. & 0.36 & 1.04 & 1.25 \\
 & $S \rightarrow T$ Comp. & 0.35 & 1.02 & 1.19 \\
 & $T \rightarrow S$ Decomp. & 0.46 & 1.02 & 1.29 \\
 & $S \rightarrow T$ Decomp. & 0.44 & 1.04 & 1.29 \\ \cmidrule(l){1-5}
\multirow{5}{*}{WikiAuto (S)} & Direct & -0.06 & 1.02 & 1.02 \\
 & $T \rightarrow S$ Comp. & 0.26 & 1.05 & 1.22 \\
 & $S \rightarrow T$ Comp. & 0.34 & 1.03 & 1.32 \\
 & $T \rightarrow S$ Decomp. & \textbf{0.81} & 1.00 & 1.31 \\
 & $S \rightarrow T$ Decomp. & 0.51 & 1.10 & 1.40 \\ \cmidrule(l){1-5}
\multirow{5}{*}{WikiAuto (T)} & Direct & 0.55 & 1.12 & 1.10 \\
 & $T \rightarrow S$ Comp. & 0.66 & 1.10 & 1.43 \\
 & $S \rightarrow T$ Comp. & 0.85 & 1.08 & 1.53 \\
 & $T \rightarrow S$ Decomp. & 0.80 & 1.21 & 1.12 \\
 & $S \rightarrow T$ Decomp. & 0.75 & 1.18 & 1.17 \\ \cmidrule(l){1-5}
\multirow{5}{*}{CLEAR (S)} & Direct & 0.31 & 1.04 & 1.09 \\
 & $T \rightarrow S$ Comp. & 0.43 & 1.15 & 1.49 \\
 & $S \rightarrow T$ Comp. & 0.34 & 1.03 & 1.18 \\
 & $T \rightarrow S$ Decomp. & 0.47 & 1.18 & 1.66 \\
 & $S \rightarrow T$ Decomp. & 0.44 & 1.09 & 1.41 \\ \cmidrule(l){1-5}
\multirow{5}{*}{CLEAR (T)} & Direct & 0.46 & 1.06 & 1.12 \\
 & $T \rightarrow S$ Comp. & 0.67 & 1.14 & 1.37 \\
 & $S \rightarrow T$ Comp. & 0.45 & 1.09 & 1.16 \\
 & $T \rightarrow S$ Decomp. & 0.83 & 1.28 & 1.44 \\
 & $S \rightarrow T$ Decomp. & 0.62 & 1.14 & 1.24 \\ \bottomrule
\end{tabular}
\caption{Human Evaluation Averages. Bold: significantly best ($p<0.05$). \textbf{Abbreviations:} Simp.: Relative Simplicity; Add.: Added Information Scale; Rem.: Removed Information Scale; Comp.: Composition; Decomp.: Decomposition; (S): Source; (T): Translation.}
\label{tab:human_eval_small}
\end{table}

To validate these findings, we calculated Inter-Annotator Agreement using the simulation methodology of \citet{pavlick-tetreault-2016-empirical}. We employed a weighted quadratic kappa, treating one random rating as the primary reference and the rounded average of the remaining ratings as the secondary, repeating the process 1,000 times to obtain stable medians. We report the following median agreement scores and 95\% confidence intervals: 
\textbf{Relative Simplicity}: 0.216 [0.202, 0.223]; 
\textbf{Added Information Scale}: 0.192 [0.177, 0.205]; 
\textbf{Removed Information Scale}: 0.360 [0.349, 0.370].

The lower agreement on relative simplicity and added information is consistent with prior reports of lower agreement on simplicity than on meaning preservation~\cite{sulem-etal-2018-simple} and with the view that simplicity is an inherently subjective dimension whose perception varies with the reader's profile and expectations~\cite{grabar-saggion-2022-evaluation}, while the higher agreement on information removal reflects the comparatively objective nature of detecting deleted content.

\section{Linguistic Features Results}
\label{app:linguistic-features}

This appendix reports the per-corpus values of a representative subset of linguistic features (lexical richness, syntactic tree depth, number of sentences, words per sentence, short-sentence ratio, Flesch-Kincaid grade, and Flesch Reading Ease) for every model and prompting strategy, providing the detailed evidence behind the feature-based analysis discussed in the main text. Per-corpus results are presented in Table~\ref{tab:asset} (ASSET), Table~\ref{tab:wikiauto} (WikiAuto), Table~\ref{tab:cochrane} (MultiCochrane), Table~\ref{tab:clear} (CLEAR), and Table~\ref{tab:wikifr} (WikiLarge-FR).

\captionsetup{font=small}
\setlength{\tabcolsep}{4pt}
\renewcommand{\arraystretch}{0.9}




\begin{table*}[!ht]
    \centering
    \footnotesize
    \setlength{\tabcolsep}{4pt} 
    \setlength{\tabcolsep}{3pt} 
    \resizebox{\textwidth}{!}{%
    \begin{tabular}{@{}llccccccc@{}}
    \toprule
    \textbf{Model} & \textbf{Prompt} & \textbf{Lex. Rich.} $\uparrow$ & \textbf{Tree Depth} $\downarrow$ & \textbf{Sent. \#} $\uparrow$ & \textbf{Words/Sent.} $\downarrow$ & \textbf{Short S.} $\uparrow$ & \textbf{FK Grade} $\downarrow$ & \textbf{FR Ease} $\uparrow$ \\ \midrule
    \multirow{5}{*}{GPT-3.5} & Direct & 0.90 & 4.62 & 1.04 & 24.55 & 0.05 & 10.13 & 60.60 \\
     & $T \rightarrow S$ Composition & 0.92 & 4.22 & 1.05 & 20.34 & 0.12 & 8.42 & 66.75 \\
     & $S \rightarrow T$ Composition & 0.92 & 4.31 & 1.02 & 20.38 & 0.10 & 8.57 & 65.86 \\
     & $T \rightarrow S$ Decomposition & 0.91 & 4.51 & 1.14 & 21.94 & 0.08 & 8.83 & 65.99 \\
     & $S \rightarrow T$ Decomposition & 0.91 & 4.66 & \textbf{1.24} & 21.32 & 0.06 & 8.66 & 66.81 \\ \cmidrule(l){1-9}
    \multirow{5}{*}{GPT-4o mini} & Direct & 0.91 & 4.64 & 1.02 & 23.50 & 0.05 & 9.51 & 63.08 \\
     & $T \rightarrow S$ Composition & 0.92 & \textbf{4.01} & 1.03 & \textbf{18.59} & 0.12 & \textbf{7.27} & \textbf{71.92} \\
     & $S \rightarrow T$ Composition & 0.92 & 4.28 & 1.02 & 20.77 & 0.10 & 8.40 & 67.45 \\
     & $T \rightarrow S$ Decomposition & 0.92 & 4.35 & 1.01 & 20.69 & 0.08 & 8.23 & 68.14 \\
     & $S \rightarrow T$ Decomposition & 0.91 & 4.45 & 1.01 & 21.68 & 0.07 & 8.81 & 65.87 \\ \cmidrule(l){1-9}
    \multirow{5}{*}{Gemini Lite} & Direct & 0.92 & 4.37 & 1.03 & 22.02 & 0.06 & 9.01 & 64.26 \\
     & $T \rightarrow S$ Composition & 0.93 & 3.82 & \textbf{1.18} & 16.79 & 0.16 & 6.58 & 73.74 \\
     & $S \rightarrow T$ Composition & 0.94 & 3.81 & 1.06 & 16.45 & 0.21 & 6.92 & 71.11 \\
     & $T \rightarrow S$ Decomposition & 0.94 & 3.73 & 1.05 & 16.97 & 0.20 & 6.77 & 72.69 \\
     & $S \rightarrow T$ Decomposition & 0.93 & 3.90 & 1.01 & 17.52 & 0.20 & 7.67 & 67.24 \\ \cmidrule(l){1-9}
    \multirow{5}{*}{DeepSeek} & Direct & 0.92 & 4.33 & 1.01 & 21.83 & 0.07 & 9.19 & 62.96 \\
     & $T \rightarrow S$ Composition & 0.89 & 4.28 & \textbf{1.24} & 19.96 & 0.09 & 8.33 & 66.34 \\
     & $S \rightarrow T$ Composition & 0.92 & 4.31 & 1.01 & 21.71 & 0.06 & 9.27 & 62.30 \\
     & $T \rightarrow S$ Decomposition & 0.93 & 4.10 & 1.01 & 19.46 & 0.10 & 8.01 & 67.81 \\
     & $S \rightarrow T$ Decomposition & 0.93 & 4.07 & 1.00 & 18.76 & \textbf{0.15} & 8.16 & 66.21 \\ \cmidrule(l){1-9}
    \multirow{5}{*}{Aya101} & Direct & 0.90 & 4.59 & 1.01 & 23.94 & 0.07 & 9.64 & 63.03 \\
     & $T \rightarrow S$ Composition & 0.91 & 4.60 & 1.00 & 24.26 & 0.06 & 9.84 & 62.19 \\
     & $S \rightarrow T$ Composition & 0.90 & 4.64 & 1.01 & 24.05 & 0.06 & 9.73 & 62.48 \\
     & $T \rightarrow S$ Decomposition & 0.90 & 4.59 & 1.02 & 23.68 & 0.07 & 9.65 & 62.73 \\
     & $S \rightarrow T$ Decomposition & 0.91 & \textbf{4.33} & \textbf{1.05} & \textbf{21.09} & 0.07 & \textbf{8.70} & 65.47 \\ \cmidrule(l){1-9}
    \multirow{5}{*}{Mistral-NeMo} & Direct & 0.92 & 4.15 & 1.01 & 21.23 & 0.10 & 9.04 & 63.24 \\
     & $T \rightarrow S$ Composition & 0.90 & 4.56 & 1.00 & 23.39 & 0.07 & 9.71 & 62.02 \\
     & $S \rightarrow T$ Composition & 0.90 & 4.65 & 1.00 & 24.23 & 0.06 & 9.99 & 61.05 \\
     & $T \rightarrow S$ Decomposition & 0.93 & 4.10 & 1.01 & 19.53 & 0.17 & 8.33 & 65.97 \\
     & $S \rightarrow T$ Decomposition & 0.93 & \textbf{3.85} & 1.01 & \textbf{16.97} & 0.22 & \textbf{7.37} & 68.76 \\ \cmidrule(l){1-9}
    \multirow{5}{*}{GPT-OSS-20B} & Direct & 0.92 & 4.18 & 1.09 & 18.54 & 0.10 & 8.80 & 63.02 \\
     & $T \rightarrow S$ Composition & 0.93 & \textbf{3.88} & 1.10 & \textbf{15.76} & 0.18 & \textbf{7.52} & 67.61 \\
     & $S \rightarrow T$ Composition & 0.92 & 4.12 & 1.12 & 17.65 & 0.14 & 8.23 & 65.69 \\
     & $T \rightarrow S$ Decomposition & 0.92 & 4.10 & 1.14 & 17.54 & 0.12 & 8.06 & 66.67 \\
     & $S \rightarrow T$ Decomposition & 0.92 & 4.28 & 1.10 & 18.40 & 0.11 & 8.56 & 64.55 \\ \cmidrule(l){1-9}
    \multirow{5}{*}{GPT-5.4-Nano} & Direct & 0.91 & 4.37 & 1.07 & 20.49 & 0.04 & 9.69 & 59.76 \\
     & $T \rightarrow S$ Composition & 0.93 & 4.06 & 1.09 & 18.03 & 0.07 & 8.44 & 64.49 \\
     & $S \rightarrow T$ Composition & 0.91 & 4.39 & 1.04 & 20.65 & 0.06 & 9.65 & 60.46 \\
     & $T \rightarrow S$ Decomposition & 0.93 & 4.13 & 1.08 & 18.68 & 0.08 & 8.91 & 62.40 \\
     & $S \rightarrow T$ Decomposition & 0.92 & 4.30 & 1.04 & 20.10 & 0.08 & 9.54 & 60.15 \\ \bottomrule
    \end{tabular}%
    }
    \caption{Results of selected features for the \textbf{ASSET} Corpus. Bold: significantly simpler than each other strategy for the same model and feature ($p<0.05$, Welch's $t$-test).}
    Arrows ($\uparrow, \downarrow$) indicate the direction of improved simplicity for each metric.

    \footnotesize\textbf{Abbreviations:}  Lex. Rich.: Lexical Richness; Tree Depth: Syntactic Tree Depth; Sent. \#: Number of Sentences; Words/Sent.: Average Words per Sentence; Short S. Ratio: Ratio of Short Sentences; FK Grade: Flesch Kincaid Grade Level; FR Ease: Flesch Reading Ease.
    \label{tab:asset}
    \end{table*}

    \begin{table*}[!ht]
    \centering
    \footnotesize
    \setlength{\tabcolsep}{3pt} 
    \resizebox{\textwidth}{!}{%
    \begin{tabular}{@{}llccccccc@{}}
    \toprule
    \textbf{Model} & \textbf{Prompt} & \textbf{Lex. Rich.} $\uparrow$ & \textbf{Tree Depth} $\downarrow$ & \textbf{Sent. \#} $\uparrow$ & \textbf{Words/Sent.} $\downarrow$ & \textbf{Short S.} $\uparrow$ & \textbf{FK Grade} $\downarrow$ & \textbf{FR Ease} $\uparrow$ \\ \midrule
    \multirow{5}{*}{GPT-3.5} & Direct & 0.90 & 4.68 & 1.10 & 25.15 & 0.06 & 9.87 & 63.26 \\
     & $T \rightarrow S$ Composition & \textbf{0.92} & 4.25 & 1.09 & 20.96 & \textbf{0.12} & 8.39 & 67.61 \\
     & $S \rightarrow T$ Composition & 0.91 & 4.26 & 1.07 & 20.82 & 0.11 & 8.27 & 68.71 \\
     & $T \rightarrow S$ Decomposition & 0.90 & 4.45 & 1.27 & 21.64 & 0.09 & 8.45 & 68.21 \\
     & $S \rightarrow T$ Decomposition & 0.89 & 4.56 & \textbf{1.39} & 20.64 & 0.09 & 8.23 & 68.43 \\ \cmidrule(l){1-9}
    \multirow{5}{*}{GPT-4o mini} & Direct & 0.90 & 4.48 & 1.04 & 25.03 & 0.06 & 9.51 & 64.80 \\
     & $T \rightarrow S$ Composition & \textbf{0.93} & \textbf{3.89} & \textbf{1.08} & \textbf{18.89} & \textbf{0.13} & \textbf{6.98} & \textbf{73.90} \\
     & $S \rightarrow T$ Composition & 0.91 & 4.25 & 1.07 & 21.74 & 0.09 & 8.24 & 69.47 \\
     & $T \rightarrow S$ Decomposition & 0.92 & 4.16 & 1.05 & 21.74 & 0.08 & 8.11 & 69.87 \\
     & $S \rightarrow T$ Decomposition & 0.91 & 4.36 & 1.04 & 22.66 & 0.08 & 8.74 & 67.27 \\ \cmidrule(l){1-9}
    \multirow{5}{*}{Gemini Lite} & Direct & 0.91 & 4.28 & 1.06 & 22.74 & 0.09 & 8.85 & 66.04 \\
     & $T \rightarrow S$ Composition & 0.93 & 3.70 & \textbf{1.29} & 16.57 & 0.21 & \textbf{6.10} & \textbf{76.39} \\
     & $S \rightarrow T$ Composition & 0.93 & 3.69 & 1.11 & 16.85 & 0.22 & 6.77 & 72.56 \\
     & $T \rightarrow S$ Decomposition & \textbf{0.94} & 3.66 & 1.08 & 17.17 & 0.20 & 6.57 & 73.87 \\
     & $S \rightarrow T$ Decomposition & 0.93 & 3.85 & 1.04 & 18.26 & 0.18 & 7.37 & 70.20 \\ \cmidrule(l){1-9}
    \multirow{5}{*}{DeepSeek} & Direct & 0.92 & 4.32 & 1.06 & 22.76 & 0.07 & 8.99 & 65.29 \\
     & $T \rightarrow S$ Composition & 0.90 & 4.23 & \textbf{1.31} & 20.25 & 0.10 & 7.92 & 69.05 \\
     & $S \rightarrow T$ Composition & 0.92 & 4.35 & 1.05 & 22.89 & 0.07 & 9.09 & 64.84 \\
     & $T \rightarrow S$ Decomposition & \textbf{0.93} & \textbf{4.03} & 1.06 & 19.91 & 0.11 & \textbf{7.76} & \textbf{69.78} \\
     & $S \rightarrow T$ Decomposition & 0.92 & 4.09 & 1.02 & 19.94 & \textbf{0.13} & 8.10 & 68.16 \\ \cmidrule(l){1-9}
    \multirow{5}{*}{Aya101} & Direct & 0.89 & 4.54 & 1.04 & 25.07 & 0.06 & 9.50 & 64.98 \\
     & $T \rightarrow S$ Composition & 0.89 & 4.59 & 1.01 & 26.36 & 0.06 & 9.89 & 63.92 \\
     & $S \rightarrow T$ Composition & 0.89 & 4.58 & 1.02 & 25.94 & 0.06 & 9.81 & 64.10 \\
     & $T \rightarrow S$ Decomposition & 0.89 & 4.55 & 1.04 & 25.41 & 0.07 & 9.62 & 64.55 \\
     & $S \rightarrow T$ Decomposition & \textbf{0.91} & \textbf{4.23} & \textbf{1.05} & \textbf{21.45} & \textbf{0.09} & \textbf{8.45} & \textbf{67.20} \\ \cmidrule(l){1-9}
    \multirow{5}{*}{Mistral-NeMo} & Direct & 0.91 & 4.23 & 1.02 & 22.31 & 0.11 & 8.94 & 65.16 \\
     & $T \rightarrow S$ Composition & 0.90 & 4.54 & 1.02 & 25.03 & 0.07 & 9.72 & 63.86 \\
     & $S \rightarrow T$ Composition & 0.89 & 4.64 & 1.02 & 26.27 & 0.06 & 10.06 & 63.04 \\
     & $T \rightarrow S$ Decomposition & 0.92 & 4.03 & \textbf{1.04} & 20.09 & 0.17 & 7.92 & 69.35 \\
     & $S \rightarrow T$ Decomposition & \textbf{0.93} & \textbf{3.73} & 1.03 & \textbf{17.54} & \textbf{0.23} & \textbf{7.16} & \textbf{70.67} \\ \cmidrule(l){1-9}
    \multirow{5}{*}{GPT-OSS-20B} & Direct & 0.91 & 4.15 & 1.19 & 18.29 & 0.11 & 8.51 & 64.92 \\
     & $T \rightarrow S$ Composition & \textbf{0.93} & \textbf{3.78} & 1.15 & \textbf{15.68} & \textbf{0.16} & \textbf{7.42} & \textbf{68.42} \\
     & $S \rightarrow T$ Composition & 0.92 & 4.05 & 1.21 & 17.12 & 0.14 & 8.07 & 66.14 \\
     & $T \rightarrow S$ Decomposition & 0.92 & 4.03 & \textbf{1.25} & 17.06 & 0.13 & 7.94 & 66.88 \\
     & $S \rightarrow T$ Decomposition & 0.91 & 4.18 & 1.19 & 18.28 & 0.12 & 8.51 & 64.85 \\ \cmidrule(l){1-9}
    \multirow{5}{*}{GPT-5.4-Nano} & Direct & 0.91 & 4.27 & 1.15 & 20.63 & 0.06 & 9.61 & 60.96 \\
     & $T \rightarrow S$ Composition & 0.92 & \textbf{3.93} & \textbf{1.18} & \textbf{17.38} & 0.09 & \textbf{8.09} & \textbf{66.43} \\
     & $S \rightarrow T$ Composition & 0.91 & 4.24 & 1.11 & 20.26 & 0.07 & 9.42 & 61.69 \\
     & $T \rightarrow S$ Decomposition & 0.92 & 4.03 & 1.13 & 18.53 & 0.08 & 8.72 & 63.80 \\
     & $S \rightarrow T$ Decomposition & 0.91 & 4.25 & 1.09 & 20.39 & 0.07 & 9.48 & 61.34 \\ \bottomrule
    \end{tabular}%
    }
    
    \caption{Results of selected features for the \textbf{WikiAuto} Corpus. Bold: significantly simpler than each other strategy for the same model and feature ($p<0.05$, Welch's $t$-test).}
    Arrows ($\uparrow, \downarrow$) indicate the direction of improved simplicity for each metric.
    
    \footnotesize\textbf{Abbreviations:}  Lex. Rich.: Lexical Richness; Tree Depth: Syntactic Tree Depth; Sent. \#: Number of Sentences; Words/Sent.: Average Words per Sentence; Short S. Ratio: Ratio of Short Sentences; FK Grade: Flesch Kincaid Grade Level; FR Ease: Flesch Reading Ease.
    \label{tab:wikiauto}
    \end{table*}

    \begin{table*}[!ht]
    \centering
    \footnotesize
    \setlength{\tabcolsep}{3pt} 
    \resizebox{\textwidth}{!}{%
    \begin{tabular}{@{}llccccccc@{}}
    \toprule
    \textbf{Model} & \textbf{Prompt} & \textbf{Lex. Rich.} $\uparrow$ & \textbf{Tree Depth} $\downarrow$ & \textbf{Sent. \#} $\uparrow$ & \textbf{Words/Sent.} $\downarrow$ & \textbf{Short S.} $\uparrow$ & \textbf{FK Grade} $\downarrow$ & \textbf{FR Ease} $\uparrow$ \\ \midrule
    \multirow{5}{*}{GPT-3.5} & Direct & 0.89 & 5.24 & 1.02 & 27.96 & 0.05 & 13.00 & 45.19 \\
     & $T \rightarrow S$ Composition & 0.89 & 5.11 & 1.01 & 25.25 & 0.08 & 11.47 & 53.39 \\
     & $S \rightarrow T$ Composition & 0.88 & 5.47 & 1.04 & 27.40 & 0.05 & 12.45 & 49.54 \\
     & $T \rightarrow S$ Decomposition & 0.89 & 5.19 & 1.03 & 24.39 & 0.08 & 11.17 & 54.44 \\
     & $S \rightarrow T$ Decomposition & 0.88 & 5.37 & 1.05 & 25.04 & 0.06 & 11.42 & 53.50 \\ \cmidrule(l){1-9}
    \multirow{5}{*}{GPT-4o mini} & Direct & 0.88 & 5.56 & 1.03 & 29.73 & 0.04 & 12.94 & 48.67 \\
     & $T \rightarrow S$ Composition & 0.91 & \textbf{4.73} & 1.03 & \textbf{22.09} & 0.08 & \textbf{9.47} & \textbf{62.62} \\
     & $S \rightarrow T$ Composition & 0.89 & 5.21 & 1.06 & 25.24 & 0.05 & 10.91 & 57.16 \\
     & $T \rightarrow S$ Decomposition & 0.89 & 5.08 & 1.04 & 25.20 & 0.07 & 10.99 & 56.42 \\
     & $S \rightarrow T$ Decomposition & 0.89 & 5.30 & 1.05 & 26.48 & 0.05 & 11.85 & 52.28 \\ \cmidrule(l){1-9}
    \multirow{5}{*}{Gemini Lite} & Direct & 0.88 & 5.39 & 1.05 & 28.13 & 0.06 & 12.67 & 48.31 \\
     & $T \rightarrow S$ Composition & 0.91 & 4.73 & 1.18 & 21.05 & 0.12 & 8.74 & \textbf{65.71} \\
     & $S \rightarrow T$ Composition & 0.91 & 4.66 & 1.09 & 20.99 & 0.12 & 9.64 & 59.64 \\
     & $T \rightarrow S$ Decomposition & 0.92 & 4.52 & 1.10 & 20.35 & 0.13 & 9.09 & 61.82 \\
     & $S \rightarrow T$ Decomposition & 0.91 & 4.72 & 1.06 & 21.06 & 0.12 & 10.06 & 56.59 \\ \cmidrule(l){1-9}
    \multirow{5}{*}{DeepSeek} & Direct & 0.90 & 5.39 & 1.03 & 26.90 & 0.05 & 12.46 & 48.25 \\
     & $T \rightarrow S$ Composition & 0.88 & 5.38 & \textbf{1.16} & 25.35 & 0.07 & 11.77 & 50.87 \\
     & $S \rightarrow T$ Composition & 0.89 & 5.32 & 1.02 & 26.89 & 0.05 & 12.56 & 47.75 \\
     & $T \rightarrow S$ Decomposition & 0.91 & 4.89 & 1.04 & 22.05 & 0.08 & 10.27 & 56.70 \\
     & $S \rightarrow T$ Decomposition & 0.91 & 4.83 & 1.01 & 21.76 & 0.09 & 10.58 & 54.82 \\ \cmidrule(l){1-9}
    \multirow{5}{*}{Aya101} & Direct & 0.86 & 5.67 & 1.02 & 30.74 & 0.04 & 11.94 & 52.44 \\
     & $T \rightarrow S$ Composition & 0.86 & 5.70 & 1.01 & 31.13 & 0.05 & 12.15 & 51.52 \\
     & $S \rightarrow T$ Composition & 0.86 & 5.53 & 1.01 & 30.75 & 0.04 & 12.06 & 51.87 \\
     & $T \rightarrow S$ Decomposition & 0.86 & 5.56 & 1.01 & 30.85 & 0.04 & 12.15 & 51.32 \\
     & $S \rightarrow T$ Decomposition & 0.86 & 5.38 & \textbf{1.06} & \textbf{27.95} & 0.06 & 11.35 & 52.97 \\ \cmidrule(l){1-9}
    \multirow{5}{*}{Mistral-NeMo} & Direct & 0.89 & 5.26 & 1.03 & 25.97 & 0.06 & 11.58 & 51.71 \\
     & $T \rightarrow S$ Composition & 0.87 & 5.70 & 1.02 & 29.88 & 0.04 & 12.94 & 48.35 \\
     & $S \rightarrow T$ Composition & 0.86 & 5.88 & 1.01 & 31.79 & 0.04 & 13.42 & 47.61 \\
     & $T \rightarrow S$ Decomposition & 0.89 & 4.96 & 1.03 & 23.63 & 0.11 & 10.61 & 56.41 \\
     & $S \rightarrow T$ Decomposition & \textbf{0.91} & \textbf{4.47} & 1.03 & \textbf{18.61} & \textbf{0.22} & \textbf{8.93} & \textbf{60.20} \\ \cmidrule(l){1-9}
    \multirow{5}{*}{GPT-OSS-20B} & Direct & 0.89 & 5.09 & 1.15 & 22.51 & 0.06 & 10.62 & 56.95 \\
     & $T \rightarrow S$ Composition & \textbf{0.92} & \textbf{4.47} & 1.08 & \textbf{18.59} & 0.12 & \textbf{8.97} & 62.25 \\
     & $S \rightarrow T$ Composition & 0.90 & 4.91 & 1.15 & 20.61 & 0.09 & 9.85 & 59.40 \\
     & $T \rightarrow S$ Decomposition & 0.90 & 4.91 & 1.17 & 21.06 & 0.09 & 9.86 & 60.00 \\
     & $S \rightarrow T$ Decomposition & 0.89 & 5.23 & 1.16 & 22.33 & 0.08 & 10.36 & 58.16 \\ \cmidrule(l){1-9}
    \multirow{5}{*}{GPT-5.4-Nano} & Direct & 0.88 & 5.35 & 1.13 & 25.18 & 0.05 & 12.24 & 49.72 \\
     & $T \rightarrow S$ Composition & 0.91 & 4.74 & 1.12 & 21.33 & 0.06 & 10.28 & 57.56 \\
     & $S \rightarrow T$ Composition & 0.89 & 5.19 & 1.07 & 25.22 & 0.04 & 12.16 & 50.46 \\
     & $T \rightarrow S$ Decomposition & 0.90 & 4.98 & 1.14 & 22.46 & 0.05 & 10.91 & 54.82 \\
     & $S \rightarrow T$ Decomposition & 0.89 & 5.24 & 1.06 & 25.22 & 0.04 & 12.19 & 50.35 \\ \bottomrule
    \end{tabular}%
    }
    \caption{Results of selected features for the \textbf{MultiCochrane} Corpus. Bold: significantly simpler than each other strategy for the same model and feature ($p<0.05$, Welch's $t$-test).}
    Arrows ($\uparrow, \downarrow$) indicate the direction of improved simplicity for each metric.
    
    \footnotesize\textbf{Abbreviations:}  Lex. Rich.: Lexical Richness; Tree Depth: Syntactic Tree Depth; Sent. \#: Number of Sentences; Words/Sent.: Average Words per Sentence; Short S. Ratio: Ratio of Short Sentences; FK Grade: Flesch Kincaid Grade Level; FR Ease: Flesch Reading Ease.
    \label{tab:cochrane}

    \end{table*}

    \begin{table*}[!ht]
    \centering
    \footnotesize
    \setlength{\tabcolsep}{3pt} 
    \resizebox{\textwidth}{!}{%
    \begin{tabular}{@{}llccccccc@{}}
    \toprule
    \textbf{Model} & \textbf{Prompt} & \textbf{Lex. Rich.} $\uparrow$ & \textbf{Tree Depth} $\downarrow$ & \textbf{Sent. \#} $\uparrow$ & \textbf{Words/Sent.} $\downarrow$ & \textbf{Short S.} $\uparrow$ & \textbf{FK Grade} $\downarrow$ & \textbf{FR Ease} $\uparrow$ \\ \midrule
    
    \multirow{5}{*}{GPT-3.5} & Direct & 0.93 & 6.30 & 1.05 & 20.66 & 0.10 & 11.77 & 42.83 \\
     & $T \rightarrow S$ Composition & 0.95 & 5.82 & 1.04 & 18.77 & 0.10 & 10.06 & 53.12 \\
     & $S \rightarrow T$ Composition & 0.93 & 6.23 & 1.02 & 20.30 & 0.14 & 11.95 & 41.00 \\
     & $T \rightarrow S$ Decomposition & 0.94 & 5.99 & \textbf{1.10} & 18.87 & 0.09 & \textbf{9.21} & \textbf{58.87} \\
     & $S \rightarrow T$ Decomposition & 0.93 & 6.34 & 1.06 & 20.14 & 0.10 & 11.08 & 47.23 \\ \cmidrule(l){1-9}
    \multirow{5}{*}{GPT-4o mini} & Direct & 0.94 & 6.15 & 1.01 & 20.43 & 0.11 & 11.92 & 41.21 \\
     & $T \rightarrow S$ Composition & 0.96 & 5.46 & 1.01 & \textbf{17.31} & 0.15 & \textbf{9.44} & \textbf{54.53} \\
     & $S \rightarrow T$ Composition & 0.94 & 5.95 & 1.01 & 20.04 & 0.10 & 11.27 & 45.25 \\
     & $T \rightarrow S$ Decomposition & 0.95 & 5.67 & 1.01 & 18.58 & 0.13 & 10.19 & 50.95 \\
     & $S \rightarrow T$ Decomposition & 0.94 & 5.82 & 1.01 & 18.69 & 0.12 & 11.09 & 44.93 \\ \cmidrule(l){1-9}
    \multirow{5}{*}{Gemini Lite} & Direct & 0.94 & 5.76 & 1.01 & 19.35 & 0.13 & 11.68 & 40.79 \\
     & $T \rightarrow S$ Composition & 0.96 & 5.20 & \textbf{1.08} & 16.52 & 0.22 & 8.80 & 57.53 \\
     & $S \rightarrow T$ Composition & 0.96 & 5.13 & 1.03 & 16.35 & 0.21 & 10.68 & 43.54 \\
     & $T \rightarrow S$ Decomposition & \textbf{0.97} & \textbf{4.73} & 1.02 & \textbf{15.01} & \textbf{0.29} & 8.64 & 56.08 \\
     & $S \rightarrow T$ Decomposition & 0.95 & 5.48 & 1.01 & 17.67 & 0.19 & 11.46 & 39.76 \\ \cmidrule(l){1-9}
    \multirow{5}{*}{DeepSeek} & Direct & 0.95 & 5.91 & 1.00 & 19.72 & 0.12 & 12.09 & 38.81 \\
     & $T \rightarrow S$ Composition & 0.95 & 5.79 & \textbf{1.03} & 19.14 & 0.12 & 11.44 & 42.72 \\
     & $S \rightarrow T$ Composition & 0.95 & 5.80 & 1.00 & 19.35 & 0.13 & 12.00 & 38.85 \\
     & $T \rightarrow S$ Decomposition & \textbf{0.97} & \textbf{4.82} & 1.00 & \textbf{15.10} & \textbf{0.27} & \textbf{9.53} & \textbf{50.31} \\
     & $S \rightarrow T$ Decomposition & 0.95 & 5.62 & 1.00 & 18.18 & 0.16 & 11.73 & 38.99 \\ \cmidrule(l){1-9}
    \multirow{5}{*}{Aya101} & Direct & 0.91 & 6.03 & 1.03 & 20.14 & 0.13 & 11.60 & 42.92 \\
     & $T \rightarrow S$ Composition & 0.92 & 6.29 & 1.02 & 21.21 & 0.10 & 12.12 & 41.16 \\
     & $S \rightarrow T$ Composition & 0.92 & 6.21 & 1.04 & 20.64 & 0.11 & 12.04 & 40.75 \\
     & $T \rightarrow S$ Decomposition & 0.91 & 6.18 & 1.01 & 20.71 & 0.11 & 11.93 & 41.51 \\
     & $S \rightarrow T$ Decomposition & 0.90 & 5.88 & \textbf{1.19} & \textbf{18.24} & 0.16 & \textbf{10.60} & \textbf{47.54} \\ \cmidrule(l){1-9}
    \multirow{5}{*}{Mistral-NeMo} & Direct & 0.96 & 5.50 & 1.01 & 17.42 & 0.19 & 11.06 & 42.68 \\
     & $T \rightarrow S$ Composition & 0.95 & 5.73 & 1.01 & 18.94 & 0.17 & 11.55 & 41.73 \\
     & $S \rightarrow T$ Composition & 0.94 & 6.01 & 1.01 & 20.12 & 0.15 & 12.36 & 37.78 \\
     & $T \rightarrow S$ Decomposition & \textbf{0.98} & \textbf{4.34} & 1.01 & \textbf{12.58} & \textbf{0.41} & \textbf{8.88} & \textbf{50.70} \\
     & $S \rightarrow T$ Decomposition & 0.96 & 5.47 & 1.01 & 17.28 & 0.22 & 11.45 & 39.79 \\ \cmidrule(l){1-9}
    \multirow{5}{*}{GPT-OSS-20B} & Direct & 0.95 & 5.40 & 1.04 & 16.29 & 0.16 & 11.38 & 41.56 \\
     & $T \rightarrow S$ Composition & \textbf{0.97} & \textbf{4.66} & 1.05 & \textbf{13.19} & \textbf{0.27} & 9.89 & 46.98 \\
     & $S \rightarrow T$ Composition & 0.95 & 5.30 & 1.05 & 15.05 & 0.18 & 11.52 & 38.49 \\
     & $T \rightarrow S$ Decomposition & 0.96 & 5.13 & 1.05 & 15.37 & 0.18 & 10.15 & 48.82 \\
     & $S \rightarrow T$ Decomposition & 0.94 & 5.33 & 1.07 & 15.62 & 0.17 & 11.59 & 39.04 \\ \cmidrule(l){1-9}
    \multirow{5}{*}{GPT-5.4-Nano} & Direct & 0.94 & 5.95 & 1.04 & 18.09 & 0.11 & 12.92 & 33.33 \\
     & $T \rightarrow S$ Composition & 0.96 & 5.13 & 1.04 & 15.81 & 0.15 & 11.07 & 42.69 \\
     & $S \rightarrow T$ Composition & 0.93 & 6.12 & 1.04 & 18.34 & 0.09 & 12.92 & 33.99 \\
     & $T \rightarrow S$ Decomposition & 0.96 & 5.27 & 1.04 & 16.30 & 0.12 & 11.36 & 41.37 \\
     & $S \rightarrow T$ Decomposition & 0.94 & 5.76 & 1.04 & 17.64 & 0.12 & 12.90 & 33.00 \\ \bottomrule
    \end{tabular}%
    }
    \caption{Results of selected features for the \textbf{CLEAR} Corpus. Bold: significantly simpler than each other strategy for the same model and feature ($p<0.05$, Welch's $t$-test).} \label{tab:clear}
     Arrows ($\uparrow, \downarrow$) indicate the direction of improved simplicity for each metric.
    
    \footnotesize\textbf{Abbreviations:} Lex. Rich.: Lexical Richness;  Tree Depth: Syntactic Tree Depth; Sent. \#: Number of Sentences; Words/Sent.: Average Words per Sentence; Short S. Ratio: Ratio of Short Sentences; FK Grade: Flesch Kincaid Grade Level; FR Ease: Flesch Reading Ease.
    \end{table*}

    \begin{table*}[!ht]
    \centering
    \footnotesize
    \setlength{\tabcolsep}{3pt} 
    \resizebox{\textwidth}{!}{%
    \begin{tabular}{@{}llccccccc@{}}
    \toprule
    \textbf{Model} & \textbf{Prompt} & \textbf{Lex. Rich.} $\uparrow$ & \textbf{Tree Depth} $\downarrow$ & \textbf{Sent. \#} $\uparrow$ & \textbf{Words/Sent.} $\downarrow$ & \textbf{Short S.} $\uparrow$ & \textbf{FK Grade} $\downarrow$ & \textbf{FR Ease} $\uparrow$ \\ \midrule
    \multirow{5}{*}{GPT-3.5} & Direct & 0.89 & 6.00 & 1.08 & 22.47 & 0.06 & 9.74 & 59.26 \\
     & $T \rightarrow S$ Composition & 0.91 & 5.53 & 1.08 & 19.46 & 0.11 & 8.23 & 66.11 \\
     & $S \rightarrow T$ Composition & 0.90 & 5.74 & 1.06 & 21.30 & 0.10 & 9.52 & 59.25 \\
     & $T \rightarrow S$ Decomposition & 0.91 & 5.54 & \textbf{1.21} & \textbf{18.54} & 0.11 & \textbf{7.53} & \textbf{69.65} \\
     & $S \rightarrow T$ Decomposition & 0.89 & 5.77 & 1.13 & 20.55 & 0.08 & 8.87 & 62.87 \\ \cmidrule(l){1-9}
    \multirow{5}{*}{GPT-4o mini} & Direct & 0.90 & 5.86 & 1.04 & 22.43 & 0.07 & 9.71 & 59.01 \\
     & $T \rightarrow S$ Composition & \textbf{0.93} & \textbf{5.20} & 1.06 & \textbf{18.28} & \textbf{0.15} & \textbf{7.64} & \textbf{68.06} \\
     & $S \rightarrow T$ Composition & 0.90 & 5.67 & 1.05 & 21.71 & 0.09 & 9.27 & 61.04 \\
     & $T \rightarrow S$ Decomposition & 0.92 & 5.60 & 1.05 & 20.18 & 0.10 & 8.44 & 64.94 \\
     & $S \rightarrow T$ Decomposition & 0.90 & 5.46 & 1.04 & 20.03 & 0.10 & 8.85 & 61.78 \\ \cmidrule(l){1-9}
    \multirow{5}{*}{Gemini Lite} & Direct & 0.91 & 5.40 & 1.07 & 20.48 & 0.11 & 9.25 & 58.87 \\
     & $T \rightarrow S$ Composition & 0.92 & 4.90 & \textbf{1.22} & 16.63 & 0.19 & 7.10 & 69.10 \\
     & $S \rightarrow T$ Composition & 0.93 & 4.92 & 1.10 & 17.41 & 0.19 & 8.01 & 63.53 \\
     & $T \rightarrow S$ Decomposition & \textbf{0.94} & \textbf{4.48} & 1.07 & \textbf{15.89} & \textbf{0.25} & 7.04 & 68.13 \\
     & $S \rightarrow T$ Decomposition & 0.92 & 5.15 & 1.05 & 18.79 & 0.14 & 8.71 & 60.24 \\ \cmidrule(l){1-9}
    \multirow{5}{*}{DeepSeek} & Direct & 0.91 & 5.70 & 1.04 & 21.52 & 0.08 & 9.65 & 57.86 \\
     & $T \rightarrow S$ Composition & 0.91 & 5.59 & \textbf{1.10} & 20.47 & 0.09 & 9.03 & 60.90 \\
     & $S \rightarrow T$ Composition & 0.91 & 5.64 & 1.03 & 21.42 & 0.09 & 9.62 & 57.93 \\
     & $T \rightarrow S$ Decomposition & \textbf{0.94} & \textbf{4.91} & 1.03 & \textbf{17.22} & \textbf{0.18} & \textbf{7.98} & \textbf{63.88} \\
     & $S \rightarrow T$ Decomposition & 0.92 & 5.42 & 1.02 & 20.15 & 0.11 & 9.29 & 58.57 \\ \cmidrule(l){1-9}
    \multirow{5}{*}{Aya101} & Direct & 0.88 & 5.96 & 1.03 & 23.17 & 0.07 & 9.96 & 58.67 \\
     & $T \rightarrow S$ Composition & 0.88 & 6.02 & 1.02 & 23.82 & 0.07 & 10.19 & 57.88 \\
     & $S \rightarrow T$ Composition & 0.88 & 5.92 & 1.02 & 23.51 & 0.07 & 10.07 & 58.23 \\
     & $T \rightarrow S$ Decomposition & 0.88 & 5.94 & 1.04 & 22.83 & 0.08 & 9.86 & 58.75 \\
     & $S \rightarrow T$ Decomposition & 0.89 & 5.84 & \textbf{1.08} & \textbf{21.18} & 0.09 & \textbf{9.35} & 60.24 \\ \cmidrule(l){1-9}
    \multirow{5}{*}{Mistral-NeMo} & Direct & 0.92 & 5.32 & 1.04 & 19.40 & 0.13 & 8.97 & 59.55 \\
     & $T \rightarrow S$ Composition & 0.91 & 5.62 & 1.04 & 20.79 & 0.10 & 9.40 & 58.67 \\
     & $S \rightarrow T$ Composition & 0.89 & 5.85 & 1.03 & 22.76 & 0.08 & 10.03 & 57.17 \\
     & $T \rightarrow S$ Decomposition & \textbf{0.95} & \textbf{4.46} & 1.04 & \textbf{15.20} & \textbf{0.29} & \textbf{7.19} & \textbf{65.72} \\
     & $S \rightarrow T$ Decomposition & 0.92 & 5.33 & 1.04 & 19.40 & 0.14 & 9.17 & 58.10 \\ \cmidrule(l){1-9}
    \multirow{5}{*}{GPT-OSS-20B} & Direct & 0.91 & 5.26 & 1.17 & 16.65 & 0.13 & 9.54 & 55.16 \\
     & $T \rightarrow S$ Composition & \textbf{0.94} & \textbf{4.66} & 1.16 & \textbf{13.82} & \textbf{0.23} & \textbf{8.04} & 61.05 \\
     & $S \rightarrow T$ Composition & 0.91 & 5.02 & 1.17 & 15.15 & 0.17 & 9.14 & 55.55 \\
     & $T \rightarrow S$ Decomposition & 0.92 & 5.05 & 1.19 & 15.55 & 0.14 & 8.58 & 60.08 \\
     & $S \rightarrow T$ Decomposition & 0.91 & 5.13 & 1.23 & 15.59 & 0.15 & 9.23 & 55.77 \\ \cmidrule(l){1-9}
    \multirow{5}{*}{GPT-5.4-Nano} & Direct & 0.90 & 5.59 & 1.12 & 18.55 & 0.08 & 10.59 & 50.48 \\
     & $T \rightarrow S$ Composition & 0.93 & \textbf{4.99} & 1.15 & \textbf{15.92} & 0.12 & \textbf{9.06} & \textbf{56.97} \\
     & $S \rightarrow T$ Composition & 0.90 & 5.29 & 1.12 & 17.36 & 0.10 & 10.11 & 52.20 \\
     & $T \rightarrow S$ Decomposition & 0.92 & 5.22 & 1.11 & 16.74 & 0.10 & 9.65 & 54.23 \\
     & $S \rightarrow T$ Decomposition & 0.91 & 5.36 & 1.09 & 17.60 & 0.11 & 10.16 & 52.32 \\ \bottomrule
    \end{tabular}%
    }
    \caption{Results of selected features for the \textbf{WikiLarge-FR} Corpus. Bold: significantly simpler than each other strategy for the same model and feature ($p<0.05$, Welch's $t$-test).}
    Arrows($\uparrow, \downarrow$) indicate the direction of improved simplicity for each metric.
    \label{tab:wikifr}
    
    \textbf{Abbreviations:} Lex. Rich.: Lexical Richness; Tree Depth: Syntactic Tree Depth; Sent. \#: Number of Sentences; Words/Sent.: Average Words per Sentence; Short S. Ratio: Ratio of Short Sentences; FK Grade: Flesch Kincaid Grade Level; FR Ease: Flesch Reading Ease.
    \end{table*}

\clearpage

\section{Error Analysis}
\label{app:error_analysis}

To clarify the qualitative behaviour of each prompting strategy and to reconcile divergent conclusions across our evaluation methods, we conducted a structured annotation of 500 GPT-3.5-Turbo outputs. We took the first 50 texts from each of two corpora drawn from different domains: CLEAR (French medical) and WikiLarge-FR (French Wikipedia), and annotated the output of all five prompting strategies for each text, yielding 500 annotated responses. Each output was compared against the English reference simplification provided by the corpus and labelled for the presence/absence of several error types. We focus on the four error types that occurred frequently enough to discriminate between strategies and that represent genuine output errors rather than stylistic variations; counts appear in Table~\ref{tab:error_counts} (Section~\ref{sec:error_analysis_main}). This appendix provides an illustrative example of each error type.

Three patterns emerge from Table~\ref{tab:error_counts}. First, Direct prompting produces by far the most errors overall (23), driven almost entirely by additions and a near-complete absence of simplification rather than by semantic faults. Second, the T\textrightarrow{}S variants are the cleanest, with the lowest totals and zero \emph{Failure to simplify} cases. Third, \emph{Propagated semantic error} appears only in the decomposition strategies, consistent with the structural prediction that splitting the task across two prompts removes the source-text context that would otherwise let the second step recover from a flawed intermediate.

\subsection*{Information addition}

This is the dominant error mode and the strongest discriminator between strategies. It is concentrated in Direct (14) and S\textrightarrow{}T composition (10), with the model inserting plausible but unsupported clarifications, parenthetical examples, or expansions of abbreviations.

\noindent\textbf{Example (CLEAR, Direct):}
\begin{quote}
\textbf{Reference (English simplification):} \textit{``The CTG on admission is a commonly used test which consists of a brief recording, generally lasting 20 minutes, of FCF and uterine activity \dots''} \\
\textbf{LLM output:} \textit{``Cardiotocography (CTG) at admission is a common screening test that involves a brief recording (usually for 20 minutes) of the fetal heart rate (FHR) and uterine activity. It is done when the mother arrives in the delivery room.''}
\end{quote}

The model expands ``CTG'' and ``FCF'' into their full forms, adds the descriptor ``screening test'', and appends a new sentence (``It is done when the mother arrives \dots''). None of these additions are present in the reference. This pattern is undetected by BLEU (since the added content reduces n-gram overlap modestly) but is captured by SARI's penalisation of unnecessary additions.

\subsection*{Failure to simplify}
This error captures cases where the output is essentially a translation with at most surface-level lexical substitution, leaving syntactic structure and lexical complexity largely unchanged. It is concentrated in Direct (5) and S\textrightarrow{}T composition (5) and is \emph{absent} from both T\textrightarrow{}S variants. This makes the T→S strategies the cleanest and is the most discriminating finding of the annotation, which directly supports the linguistic feature results in Appendix~\ref{app:linguistic-features}, where Direct outputs show the smallest reductions in syntactic tree depth and words-per-sentence.

\noindent\textbf{Example (CLEAR, Direct):}
\begin{quote}
\textbf{Reference:} \textit{``due to the reduced alertness induced by the presence of alcohol, attention is drawn to the risks associated with driving a vehicle or using a machine.''} \\
\textbf{LLM output:} \textit{``Because of decreased vigilance caused by alcohol consumption, attention is drawn to the risks associated with driving a vehicle or using machinery.''}
\end{quote}

Only ``vigilance'', ``decreased'', and ``machinery'' differ from a literal translation; sentence structure, register, and lexical level are unchanged. The output is fluent and faithful but would not meet accessibility needs.

\subsection*{Semantic inversion}
A small number of outputs (6 across all strategies) introduce a meaning-reversing change. These errors are high-severity because they are undetected by surface metrics: BLEU and BERTScore remain high even when the truth value of the sentence flips. The error appears mostly in Direct (4) and the composition variants (1 each), and is absent from the decomposition strategies in our sample.

\noindent\textbf{Example (CLEAR, Direct):}
\begin{quote}
\textbf{Reference:} \textit{``we found that for the treatment of pneumonia in outpatients, amoxicillin is an alternative treatment to cotrimoxazole.''} \\
\textbf{LLM output:} \textit{``For the treatment of outpatients with a urinary tract infection, amoxicillin is an alternative to cotrimoxazole.''}
\end{quote}

The clinical condition changes from \emph{pneumonia} to \emph{urinary tract infection}. The sentence remains fluent, the structure is preserved, and BERTScore would register both as semantically similar — yet the medical content is wrong. This illustrates a key limitation of automatic metrics for medical-domain CLTS.

\subsection*{Propagated semantic error}
This error appears \emph{only} in the decomposition strategies (T\textrightarrow{}S Decomp.: 2; S\textrightarrow{}T Decomp.: 2; zero in all single-prompt strategies). The structural explanation is direct: when the two operations are performed in separate prompts, the second step has no access to the original source and cannot detect or correct a flawed intermediate.

\noindent\textbf{Example (CLEAR, T\textrightarrow{}S Decomposition):}
\begin{quote}
\textbf{Reference:} \textit{``there are only a few studies reporting the evaluation criteria that interested us.''} \\
\textbf{LLM output:} \textit{``However, we were not interested in the evaluation criteria of two of the studies.''}
\end{quote}

The meaning is inverted (``a few studies report criteria we cared about'' $\rightarrow$ ``we did not care about the criteria''): the error is introduced in the intermediate English translation, and the simplification step, which no longer sees the French source, builds on this flawed intermediate and carries the error into the final output. This kind of error is the principal structural risk of the decomposition design and is essentially impossible in single-prompt strategies, where the model retains the source context.

\subsection*{Preliminary experiment: source-aware decomposition}
\label{app:source_aware}
To test whether giving the second prompt access to the source text mitigates propagated errors, we re-ran both decomposition strategies on the same 50 texts per corpus (CLEAR and WikiLarge-FR). Since GPT-3.5-Turbo is no longer available through the API, we used GPT-4o-mini with default decoding settings. For each text, the first step was run once, and its intermediate output was passed to two variants of the second step: the \emph{standard} prompt, identical to our main experiments, and a \emph{source-aware} prompt, which additionally contains the original French source text and instructs the model to preserve its meaning. Holding the intermediate output fixed isolates the effect of source access in the second step. We annotated each final output for propagated semantic errors, i.e., meaning errors introduced in the intermediate output and carried into the final output.

\begin{table}[ht]
\centering
\small
\begin{tabular}{llccc}
\toprule
Corpus & Order & Prop.\ err. & BLEU & SARI \\
\midrule
CLEAR & T$\rightarrow$S & 1 / 1 & 14.1 / 16.4 & 35.2 / 34.4 \\
CLEAR & S$\rightarrow$T & 1 / 1 & 20.1 / 22.7 & 36.9 / 38.1 \\
WikiLarge-FR & T$\rightarrow$S & 0 / 0 & 24.8 / 27.4 & 37.3 / 38.0 \\
WikiLarge-FR & S$\rightarrow$T & 5 / 5 & 29.5 / 30.0 & 38.0 / 38.2 \\
\bottomrule
\end{tabular}
\caption{Standard / source-aware second step for GPT-4o-mini decomposition (50 texts per corpus). Prop.\ err.: number of outputs with a propagated semantic error. No BLEU or SARI difference is significant (Welch's $t$-test, $p>0.25$).}
\label{tab:source_aware}
\end{table}

Table~\ref{tab:source_aware} shows that access to the source did not correct a single propagated error: in all 7 cases, the source-aware output reproduces the error of the intermediate step, e.g., \emph{seule une étude} (``only one study'') simplified to \emph{une étude} (``a study''), or \emph{a rampé} (``crawled'') rewritten as \emph{a ramé} (``rowed''), both carried into the English output. When the source-aware output did differ, the change typically restored content omitted by the intermediate step (e.g., confidence intervals, a dropped hedge, or a parenthetical term), which accounts for the consistent but non-significant BLEU gains. Most propagated errors (6 of 7) arise in S$\rightarrow$T decomposition, where the first step rewrites the French text before translation. These results suggest that exposing the source, even with an explicit instruction to preserve its meaning, is not enough: the model largely builds on the intermediate output rather than checking it against the source.

\subsection*{Reconciling the error analysis with the automatic and human evaluations}
The annotation makes the divergent metric results coherent: Direct's high BLEU and semantic similarity reflect its low rate of semantic errors and high rate of source-preserving outputs (\emph{Failure to simplify}), while its low SARI and weak readability improvements reflect the same conservatism. T\textrightarrow{}S approaches have both the lowest total error count and zero \emph{Failure to simplify} cases, supporting the linguistic-feature finding that they produce the most genuinely simplified outputs. The decomposition strategies' moderate error counts are dominated by a structurally specific failure (\emph{Propagated semantic error}) that no single-prompt strategy exhibits — a trade-off intrinsic to the design rather than a model deficiency.

\section{Software and Python Packages}
\label{sec:software_appendix}

To ensure full reproducibility of our experimental pipeline, evaluation metrics, and statistical analyses, we provide the exact versions and sources of the Python packages used. All experiments were conducted in a Python 3.11 environment.

\begin{itemize}
    \item \texttt{bert-score==0.3.13} \cite{zhang2020bertscoreevaluatingtextgeneration}
    \item \texttt{deep-translator==1.11.4} (\url{https://github.com/nidhaloff/deep-translator})
    \item \texttt{easse==0.2.4} \cite{alva-manchego-etal-2019-easse}
    \item \texttt{nltk==3.9.1} \cite{NLTK}
    \item \texttt{numpy==1.26.4} \cite{numpy}
    \item \texttt{openai==1.50.2} (\url{https://github.com/openai/openai-python})
    \item \texttt{openpyxl==3.1.2} (\url{https://openpyxl.readthedocs.io})
    \item \texttt{pandas==2.2.2} \cite{mckinney-proc-scipy-2010}
    \item \texttt{pyphen==0.15.0} (\url{https://github.com/Kozea/Pyphen})
    \item \texttt{python-dotenv==1.0.1} (\url{https://github.com/theskumar/python-dotenv})
    \item \texttt{scipy==1.13.0} \cite{SciPy}
    \item \texttt{sentence-transformers==3.0.1} \cite{reimers-2020-multilingual-sentence-bert}
    \item \texttt{simpletransformers==0.70.1} \cite{10.1145/3673791.3698412}
    \item \texttt{spacy==3.7.5} \cite{spacy}
    \item \texttt{textstat==0.7.3} (\url{https://github.com/shivam5992/textstat})
    \item \texttt{torch==2.4.0} \cite{10.5555/3454287.3455008}
    \item \texttt{xlsxwriter==3.2.0} (\url{https://xlsxwriter.readthedocs.io})
\end{itemize}

\end{document}